\documentclass{article}

\PassOptionsToPackage{numbers,square}{natbib}

    \usepackage[preprint]{neurips_2024}

\usepackage[utf8]{inputenc} 
\usepackage[T1]{fontenc}    
\usepackage{hyperref}       
\usepackage{url}            
\usepackage{booktabs}       
\usepackage{amsfonts}       
\usepackage{nicefrac}       
\usepackage{microtype}      
\usepackage{xcolor}         
\usepackage{enumitem}
\usepackage{subcaption} 
\usepackage{bm}
\usepackage{graphicx}
\usepackage{amsmath}
\usepackage{amsfonts}
\begin{document}

\title{Multi-Type Point Cloud Autoencoder: A Complete Equivariant Embedding for Molecule Conformation and Pose}

%

\author{Michael Kilgour\\
Department of Chemistry, New York University\\
\texttt{michael.kilgour@nyu.edu}\\ 
\And
Mark E. Tuckerman\\
Courant Institute of Mathematical Sciences, New York University\\
NYU-ECNU Center for Computational Chemistry at NYU Shanghai\\
Simons Center for Computational Physical Chemistry at New York University\\
\texttt{mark.tuckerman@nyu.edu}
\And
Jutta Rogal\\
Initiative for Computational Catalysis, Flatiron Institute\\  
Department of Chemistry, New York University\\
Fachbereich Physik, Freie Universität Berlin\\
\texttt{jrogal@flatironinstiute.org}
}

\maketitle

\begin{abstract}
Representations are a foundational component of any modelling protocol, including on molecules and molecular solids.
For tasks that depend on knowledge of both molecular conformation and 3D orientation, such as the modelling of molecular dimers, clusters, or condensed phases, we desire a rotatable representation that is provably complete in the types and positions of atomic nuclei and roto-inversion equivariant with respect to the input point cloud. 
In this paper, we develop, train, and evaluate a new type of autoencoder, molecular O(3) encoding net (Mo3ENet), for multi-type point clouds, for which we propose a new reconstruction loss, capitalizing on a Gaussian mixture representation of the input and output point clouds. 
Mo3ENet is end-to-end equivariant, meaning the learned representation can be manipulated on O(3), a practical bonus. 
An appropriately trained Mo3ENet latent space comprises a universal embedding for scalar and vector molecule property prediction tasks, as well as other downstream tasks incorporating the 3D molecular pose, and we demonstrate its fitness on several such tasks.
\end{abstract}

\section{\label{sec:Introduction}Introduction}
The question of how to represent a molecule to a learning model is the crucial first step in any modelling routine.
A representation must be sufficiently rich to capture the relevant features, yet not too complex, which leads to overfitting.
3D molecular structures are typically cast as point clouds, i.e., lists of coordinates and atom types, which, by necessity, have varying sizes.
Despite extensive prior work in both point cloud modelling and molecular representations, no extant model explicitly decomposes 3D molecular structures to a fixed-dimensional equivariant embedding (not dependent on the input graph size), and then reconstructs them in atomistic detail without any scaffolding, sampling, or other external information, as would, for example, an autoencoder model.

Such an autoencoder would be a useful tool in several application areas.
The perhaps most exciting of these is in the growing field of machine learning for molecular materials design~\cite{zugec2024, kilgour2023,Carpenter2021,Egorova2020,Cersonsky2023,Tom2020,Defever2019,Kadan2023,Kohler2023,Apostolakis2001}, where one may need to condition a downstream task on the precise geometry \textit{and} rotational orientation of an input molecule, represented by an embedding vector prepared by the encoder.
Further, such an encoder represents a complete, if brute-force, solution to the molecule representation learning problem.

Currently, it is common practice to co-train molecule representation and property prediction models~\cite{guo2022graph,wieder2020compact} with graph neural networks, although there is also impressive recent work on extensively pretrained representation models~\cite{Zhou2023,fang2022geometry,zhu2022unified,liu2021pre,ni2024pre,zaidi2022pre}.
Co-training is generally an efficient approach, allowing one to make good use of finite datasets.
However, in the limit of large datasets and a well-converged encoder model, we can obviate the need for graph prediction models, with their attending computational and theoretical complexities~\cite{zhang2023expressive,corso2024graph}, and train simpler models such as multilayer perceptrons directly on a pretrained embedding.

In this work, we advance the fields of point cloud learning and molecule representation by developing a new reconstruction loss and model architecture for multi-type point clouds which is smooth and fast to compute, returns trainable gradients, and allows for varying input composition and size.
We leverage developments in equivariant graph learning for point-structured data \cite{Schutt2021, Le2022,Deng2022} to develop and optimize an end-to-end equivariant autoencoder model.
If well-converged, the latent O(3) equivariant embedding between encoder and decoder in this model must necessarily contain the complete 3D coordinate and type information about a given molecule, and therefore serve as a universal embedding space for molecular data. 
Such an embedding could be leveraged for any downstream task dependent on molecular structure \textit{and/or} pose, without fear of common issues in graph learning such as lack of representational power~\cite{Pozdnyakov2022}, with the added benefit of rotatability. 

Although we do not explore the possibility here, the decoder half of this autoencoder could be developed into a new generative model (variational autoencoder) for atomistic data.
Property-driven sampling and search would then be possible via by conditional training and careful engineering of the embedding space.
In the approach presented here, the decoder has the additional advantage of sampling in a single forward pass of the model, as opposed to modern diffusion approaches which may require a number of forward passes to generate a sample~\cite{Jing2022,xu2022geodiff,xu2023geometric,hoogeboom2022equivariant}.

Our Mo3ENet model comprises an equivariant graph neural network (EGNN) encoder, which embeds an input point cloud as a fixed dimensional equivariant representation $\vec{g}_{k\times3}$, with $k$ the embedding dimension, and an equivariant decoder, structured either as an EGNN or equivariant multilayer perceptron (EMLP). 
The encoder is flexible in the number of input points, and the output returns a fixed-size `swarm' of output points.
A large swarm size with learnable member weights ensures flexibility of the decoding with respect to input size and composition.
Both inputs and outputs are mapped to Gaussian mixtures (GMs), with their pairwise overlaps computed to determine the reconstruction fidelity.
Training is facilitated by an reduction of the GMs' widths until neighboring atoms no longer significantly overlap, and the distance between input and output distributions is minimized, corresponding to perfect reconstruction of the input point cloud. 

To demonstrate our autoencoder model in a chemistry context, we train it on a very broad distribution of randomly generated molecular conformers, based on QM9 and ZINC22~\cite{ruddigkeit2012enumeration,ramakrishnan2014quantum,tingle2023zinc} molecular backbones, and assess its representation ability on molecule reconstruction and property prediction tasks, as well as demonstrate that, combined with unit cell parameters, the molecule embedding can be used to predict crystal properties such as lattice energies.
This last task shows the plausibility of more complex modelling tasks, including molecular crystal search and optimization.

\section{\label{sec:Related}Related work}
\textbf{Point cloud generation}

Given their utility, there is vast prior literature on point cloud learning, including object classification, semantic segmentation, generative modelling, representation learning, and much more~\cite{Qi2017, Tchapmi2019, Yang2018, guo2020deep,zhang2019review,suchde2023point}. 
Various model architectures have been proposed to process this data type, with inputs cast as voxel grids, meshes, and real-space graphs.
Models incorporating underlying physical equivariances into their building blocks have also seen increased uptake in recent years~\cite{Li2019, Deng2022, Vignac2020, Keriven2019,finzi2021practical,liao2022equiformer}, with particular popularity in the chemical modelling community~\cite{satorras2021n,Schutt2021, Le2022}.
In this vein, several generative approaches including various autoencoders~\cite{Yang2018, Choe2022} and generative adversarial networks~\cite{Zamorski2020, Jiang2018}, have been developed, with particular focus on point cloud reconstruction.
While many insights from this field, and particularly the original PointNet architecture~\cite{ahmadi2024machine}, have been influential in molecular modelling, there are as yet limited tools for the analysis and generation of point clouds with multiple types.

\noindent\textbf{Molecule representation learning}

The study of molecule representations goes back decades in the chemistry literature.
Such approaches run the gamut from straightforward and interpretable, such as the bag of bonds~\cite{hansen2015machine} or SMILES strings, to the more structurally-informed Coulomb matrix~\cite{rupp2012fast}, to physics-inspired descriptors~\cite{bartok2013representing,musil_PhysicsInspiredStructural_2021}, and on to high-dimensional learned embeddings~\cite{Yang2019,Zhou2023,wang2022molecular,raghunathan2022molecular,liu2021pre,zhu2022unified,fang2022geometry}.

The most powerful pretrained approach currently appears to be the Uni-Mol framework~\cite{Zhou2023}, which combines a large and powerful graph model, self-supervised learning, property prediction, and very large datasets, to train a high-quality molecule representation. 
Despite impressive benchmark performance, however, Uni-Mol does not explicitly encode and decode the full molecular point cloud, instead, predicting positions and types of single corrupted or masked atoms. 

Many generative models for molecules have been developed, including a number of autoencoders~\cite{satorras2021n,You2018,Winter2021,Joo2020,Simonovsky2018,Graph2023,Ilnicka2023,Chennakesavalu2023}, though these generally either work in the space of nodes and edges (so-called `2D' structure), rather than 3D coordinates, are not generalizable between molecules, or do not encode to a single molecule-wide embedding of fixed size, without any scaffolding provided to assist reconstruction, and none return a complete and equivariant (rotatable) embedding.

\section{\label{sec:Methods}Models and methods}

\subsection{Model architecture}

\begin{figure}
\centering
    \includegraphics[width=\textwidth]{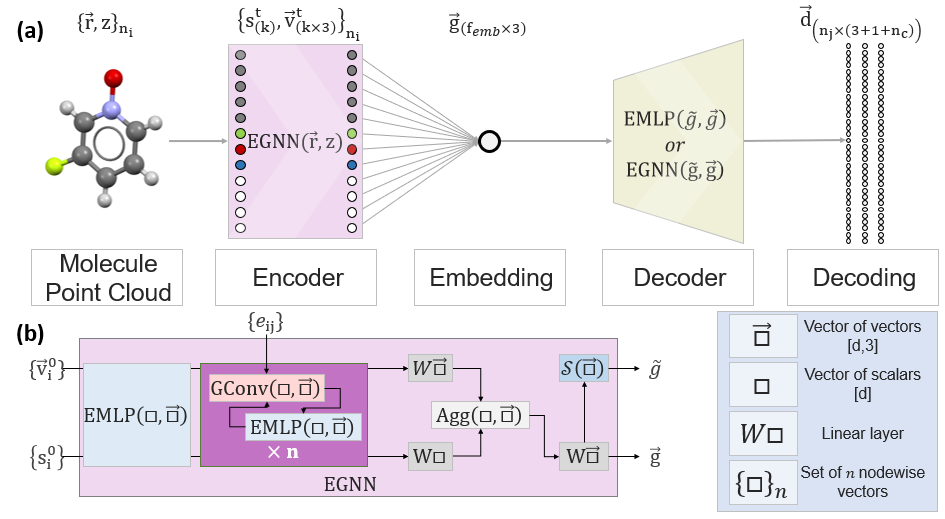}
    \caption{(a) Overall summary of autoencoder architecture.
    Molecule point cloud coordinates, $\vec{r}$ and atom types $z$ are embedded and bottlenecked to dimension $f_{emb}$ by the encoder.
    Then, the decoder (an equivariant MLP or GNN with fixed graph size) transforms the embedding to $n_j$ independently weighted points in the $3+n_c+1$ dimensional output cloud, where $n_c$ is the number of classes or types in the dataset and one extra dimension for the learned weights.
    (b) Outline of the architecture of the encoder graph model, outputting the equivariant embedding $\bm{\vec{g}}$ and its scalarized counterpart $\bm{\tilde{g}}$.}
\label{fig:architecture_summary}

\end{figure}
The Mo3ENet autoencoder is comprised of an encoder and decoder, each inputting and outputting scalars $\bm{s}$ and vectors $\bm{\vec{v}}$.
In the molecular context, the initial nodewise scalar inputs $\bm{s}$ are the atom types and radial distances from the molecule centroid, and vector inputs $\bm{\vec{v}}$ are the Cartesian point vectors from the molecule centroid, together comprising a single molecular graph.
The model is end-to-end equivariant on O(3), that is,  rotations and inversions of the input result in identical rotations to the embedding and decoding.
As illustrated in Figure \ref{fig:architecture_summary}(a), during a forward pass, the atoms in a given molecule are embedded as nodes in a graph, nonlinearly transformed, and aggregated to a fixed size latent embedding, $\bm{\vec{g}}$, which is decoded to a swarm of points $\bm{\vec{d}}$.

The primary building blocks of our model are EMLPs, graph convolution layers, and graph-wise aggregators.
In constructing these modules, we borrow concepts from prior works on equivariant models on scalars and vectors \cite{Schutt2021, Le2022, Deng2022}, which process directly the point positions in Cartesian space. 
Only a few transformations of vector features retain O(3) equivariance; we nevertheless realize a flexible model using the following basic operations: (1) weighted linear combination, 
(2) adjustment of vector lengths via multiplication by scalars, and (3) vector embedding and activation, similar to the projection scheme developed in \cite{Deng2022}, as well as the usual neural network operations for scalars.
Full technical details of model construction are provided in the appendix.

The encoder model, shown in Figure \ref{fig:architecture_summary}(b)  is comprised of an embedding step, alternating EMLP and graph convolution modules, and finally, graph aggregation. 
A `scalarizing' module $\mathcal{S}$, can also be used to generate a rich, rotation invariant scalar representation of the full vector embedding. 
The decoder can be either another equivariant graph model (sans global aggregation), or an equivariant MLP, returning scalar and vector features of the output swarm,
\begin{align}\label{eq:decoder}
\begin{split}
    \bm{s}^N, \vec{\bm{v}}^N &= \mathrm{Decoder}(\tilde{\bm{g}}, \vec{\bm{g}})\\
     \vec{\bm{d}}&= \vec{\bm{v}}^N \Vert \bm{s}^N,
\end{split}
\end{align}
with the EGNN decoder first generating scalar and vector node embeddings, and positions from the encoder embedding, for a \textit{constant} number of nodes, via a learned linear transformation.
$\vec{\bm{d}}$ is a vector with shape $n_j \times (3 + n_c + 1)$, corresponding to the $n_j$ output swarm nodes, the 3 Cartesian dimensions, the $n_c$ atom type probabilities, and the raw amplitude of the nodewise weight. 

The decoder is equivariant with respect to the Cartesian dimensions and invariant with respect to type and weight. 
The $n_c$ elements are normed via softmax within each node to a particle-type probability.
Each output point also includes a learnable weight $w_j$ computed as the softmax over the predicted amplitude for all points in the output swarm for a given molecule, multiplied by the number of atoms in the input graph, such that $\sum_j w_j = n_i$.
Note that throughout this work, $n_i$ corresponds to the number of nodes in the input molecular graph, and $n_j$ the number of nodes in the decoded output swarm.

Given a constant graph size in the decoder, the choice of EMLP or EGNN is not obviously theoretically indicated, and indeed, a graph model introduces extra complexity.
However, empirically, the graph version is observed to have superior convergence properties, and so the results in Section \ref{sec:Related} are generated using the EGNN version.
All model hyperparameters and further training details are given in the appendix. 

\subsection{Reconstruction loss}
In order to train a multi-type point cloud autoencoder, we develop a reconstruction loss function with the following properties: smoothly converging in particle positions and type variations, fast to evaluate, permutation invariant, scales easily to large numbers of particles and particle types, and provides a trainable gradient.

We cast input and output point clouds as point-centred Gaussian mixtures in the 3 Cartesian plus $n_c$ point type dimensions (corresponding practically to atom types, H, C, N, O, F,...). 
For example, a hydrogen atom with Cartesian coordinates $x, y, z$ would have coordinates $(x, y, z, 1, 0, 0, 0, 0,...)$, and a carbon atom $(x, y, z, 0, 1, 0, 0, 0,...)$.

\begin{figure}
\centering
    \includegraphics[width=1\textwidth]{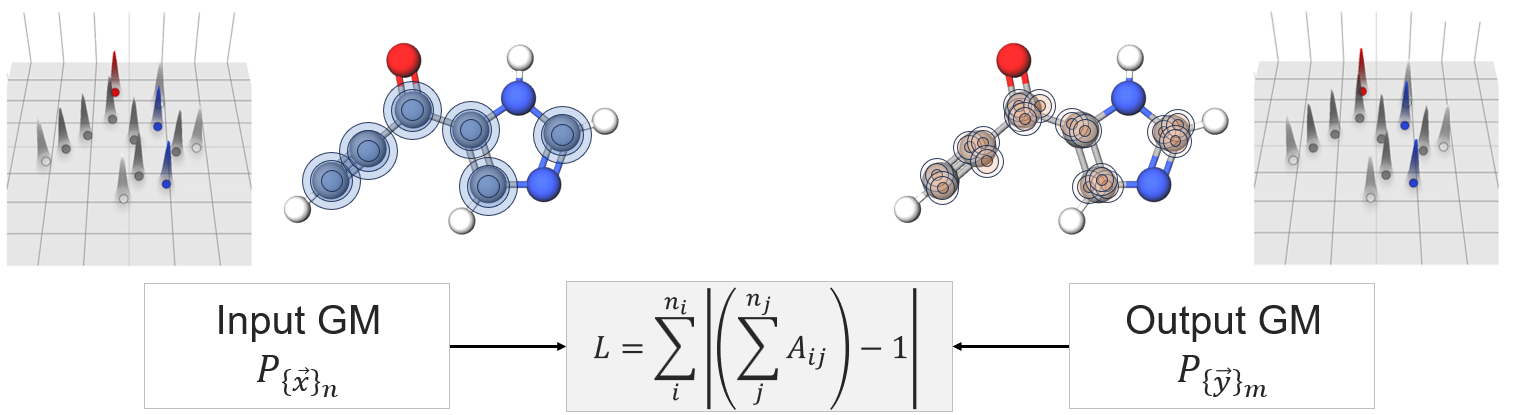}
    \caption{Illustrative diagram demonstrating the correspondence between point clouds and Gaussian Mixtures.
    We show the explicit GMs projecting in different colors each type dimension as well as cartoons of point-centered Gaussians, projected in the `carbon' dimension. 
    These GMs are compared pairwise over input and output points to determine the overall overlap with each input node.
    }
    \label{fig:loss_cartoon}
\end{figure}

In the convergent limit, diagrammed in Figure \ref{fig:loss_cartoon}, each point in the input should be exactly overlapped by the GM of the output.
The input GM is centred on the Cartesian points in the input cloud and one-hot encoded in the $n_c$ type dimensions, with a variable number of input points $n_i$.
The output GM is constructed likewise, centred in the points in the output distribution, with learned probabilities for each particle type dimension. 
The output comprises a fixed number of points $n_j$, with $n_j$ always greater than $n_i$.

The overlap between points $i$ and $j$ in the input and output, respectively, is given by their Gaussian distributed distances,
\begin{equation}\label{eq:gauss_overlap}
    A_{ij}(\vec{x}_i,\vec{y}_j)= w_j \cdot e^{-\frac{d_{ij}^2}{\sigma}},
\end{equation}
with $d_{ij}$ as the euclidean distance between points $\vec{x}_i$ and $\vec{y}_j$, $\vec{x}_i=\vec{r}_i\Vert e_i$, $\vec{r}_i$ the 3D coordinate, and $e_i=c_T\cdot \mathrm{OneHot}(z_i)$ a one-hot encoding for the type of particle $i$. 
$w_j$ is, again, the learned weight for output point $j$.
The predicted points $j$ are similarly encoded as $\vec{y}_j=\vec{r}_j\Vert\left(c_T\cdot P(z_j)\right)$, with the normalized probabilities over particle types substituted for the one-hot encoding. 
$c_T$ is a scaling factor that adjusts the relative distances between particle types.
It is a tunable hyperaparamter that can have significant impact on the training dynamics if taken too large or too small compared to $\sigma$.
In principle, we expect to see good performance when $c_T$ is taken on the order of interatomic distances, typically 2\r{A}, such that atoms of different types are neither too close, such that they overlap, or too far, such that the type space becomes too sparse.

For $\sigma \ll$ the minimum inter-particle distance, the input and output GMs are matched when the pairwise overlaps of the outputs with each input particle sum to unity. 
Therefore, the per-particle loss is simply the sum of pairwise overlaps from $j\to i$
\begin{equation}\label{eq:RLoss}
    L = \sum_i^{n_i}\left|\left(\sum_j^{n_j}A_{ij}\right) - 1\right|.
\end{equation}
This functional form attracts probability mass towards each input particle when its overlap with the outputs is less than 1, and repels it when the total overlap is greater than 1.

In practice, we initialize $\sigma$ somewhat large to ensure all swarm members receive finite gradients at the beginning of training. 
In such early stages, the loss is the difference between the input-output overlap and the input self-overlap, from adjacent points' Gaussians intruding on one-another,
\begin{equation}\label{eq:RLoss_practical}
    L = \sum_i^{n_i}\left|\left(\sum_j^{n_j}A_{ij}\right)- \left(\sum_k^{n_i}A_{ik}\right)\right|.
\end{equation}

\begin{figure*}
\centering
    \includegraphics[width=\textwidth]{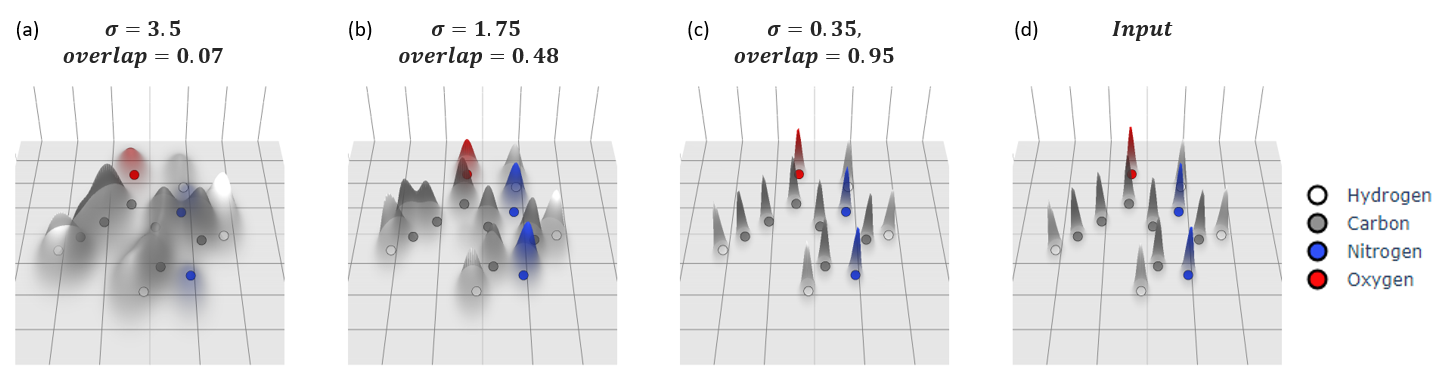}
    \caption{\label{fig:toy_gaussian} (a), (b), and (c) Convergence of the output Gaussian mixture on a flat molecule (point cloud shown), while reducing the Gaussian width, projecting each of the typewise dimensions in individual colors. (d) Input GM.}
\end{figure*}

A converging series of decoder output GMs for a given molecule is shown in Figure \ref{fig:toy_gaussian}.
As the reconstruction loss improves, $\sigma$ is reduced, increasing the problem difficulty, until the overlap between points in the input becomes negligible. 
The $\sigma$ reduction schedule, and specifically the threshold loss at which we trigger a reduction, is a critical hyperparameter affecting the rate of model convergence.
If this threshold is set low, the model is required to have nearly perfect overlap at a given $\sigma$ before narrowing, meaning the relevant Gaussian components are very close to the atoms.
Since the gradients at the peak of a Gaussian vanish, convergence in this regime can be extremely slow.
Empirically, a reconstruction loss threshold of 0.15 was found to optimize the speed of convergence, supplying sufficient gradients without stranding output GM components too far from input atoms when $\sigma$ is narrowed.

We experimented with several alternate or auxiliary losses, for example, one enforcing the correct distribution of atom types in the decoder output, with the goal of accelerating model convergence.
In practice, however, these did not generally improve model performance, possibly due to limiting the optimization pathways for the basic reconstruction loss.
Three exceptions were employed to ensure robust gradients and no `orphaned' nodes. 
\begin{equation}\label{eq:constraint1}  
    L_c=\sum_j^{n_j}\mathrm{Max}\{(\lVert\vec{r}_j\rVert - r_{mol}), 0\}
\end{equation}
constrains output points to be within the maximum radius of a given molecule.
This prevents decoded points straying too far from the range of the input, which would lead to vanishingly small gradients from the reconstruction loss.
Secondly, since we observed some point weights vanishing in early epochs and inhibiting convergence, we implement a penalty on positive deviations of the output point weights $w_j$ from the mean for a given output swarm. 
In practice, both constraining losses converge trivially and do not seem to haper optimization against the reconstruction loss.
Finally, we impose a penalty on the distance of atom $i$ to the nearest swarm member $j$, which when applied gently helps keep atoms from getting missed entirely by the swarm, say if they are too far away for the gaussian overlap loss to develop a finite gradient.
\begin{equation}\label{eq:constraint2}  
    L_d=c_d\sum_i^{n_i}\min_{j}d_{ij}^2.
\end{equation}
where $c_d$ is a hyperparameter.  
At large values of $c_d$, training is significantly slowed, so we take in practice $c_d=0.1$.

The output dimension $n_j\times (4+n_c)$ of the decoder model is typically high.
Further, the loss calculation requires an all-to-all distance calculation between atoms and swarm nodes, naively scaling quadratically in both. 
We have trained EMLP decoders on molecules with up to 1024 swarm members, 
but have not noticed bottlenecking in training from these parts of the model. 
Perhaps for very large point clouds with many atom types, one might require more clever engineering to address these issues, but at this time this does not appear to be necessary.

\subsection{Distance Calculation}
Beyond the reconstruction loss, which is somewhat abstract in chemical structure terms, we also calculate a mean distance between input and output point clouds, defined as distance between all input atoms and their partners in the decoded output (see the rigorous definition in the appendix).

Since Mo3ENet is equivariant and molecules are always centered on the origin, there is no need for overall structure matching, only assignment of output GM components to input atoms.
When given a molecular scaffold, this can be straightforwardly done by assigning any output cloud points $j$ within a short range (0.5\r{A}) of inputs $i$ to cluster $i$, and aggregating them to synthetic particle predictions by averaging according to their respective weights.
The atomwise deviations are then easily calculated and averaged molecule-wise. 
In the case of no molecular scaffold, such as during \textit{de novo} generation or conformational transformation, agglomerative clustering can be performed on the output swarm in the Cartesian dimensions, followed by a merge of low-weight clusters into their nearest neighbors. 

Both of these approaches qualitatively agree with visual inspection and the reconstruction loss (probability mass overlap between input and output GMs) in the range of small deviations, where the output swarm particles from a trained autoencoder model are generally tightly clustered around the predicted particle positions. 
When the output swarm is diffuse, meaning the reconstruction is under-converged, the reconstruction loss is large, and both clustering approaches fail.
In other words, these clustering approaches only return sensible results when applied to an already reasonably good reconstruction.

\subsection{Molecular Data}
The point clouds contained in molecular datasets such as QM9 are frequently optimized against some potential energy function, representing a ground state or zero temperature structure. 
While these conformers are `probable' structures, molecules are generally at least somewhat flexible, and a given molecule may adopt a wide range of conformations.
An autoencoder trained on only highly optimized structures may struggle to apprehend valid structures in non-ideal conformations, or conformations which are unlikely under a given potential.
Therefore, to ensure generalization to a variety of molecular conformations, we developed a pipeline for synthesizing large numbers of random conformers.

Before training, we extracted all unique canonicalized SMILES codes from the QM9 and ZINC22 datasets containing H, C, N, O, and F, up to 9 and 14 heavy atoms, respectively.
10k optimized molecular structures from QM9 were set aside for evaluation, and the remaining 7.2M SMILES are used for training.

Point clouds are generated according to the following procedure: (1) select a random SMILES code from the training set, (2) generate a rough 3D conformation with RDKit~\cite{rdkit}, (3) if the molecule has more than 9 heavy atoms, iteratively truncate it at rotatable bonds, capping with hydrogens, until it is within the appropriate range, (4) randomly spin all rotatable bonds in the molecule, (5) run very short energy minimizations to avoid severe interatomic overlaps, (6) to further boost generalization to different molecule conditions, gently noise and globally rescale the completed structure.

This dataset synthesis approach provides a very wide range of molecular structures, practically limited only by the distribution of functional groups represented in the QM9+ZINC22 SMILES, which could be amended in the future with more datasets such as GEOM~\cite{axelrod2022geom}, or QMugs~\cite{isert2022qmugs}, or by adding random generation of the molecule backbone to the above procedure.

We also attempted training on point clouds sampled from a random uniform distribution in coordinates and atom types, but found negligible benefit to model performance on molecular data.
This should, perhaps, not be surprising, since molecular information is structured by certain correlations, such as bond lengths and angles, which random point clouds lack, and therefore are far outside the relevant distribution.


\section{\label{sec:Results}Results}
\subsection{Molecule reconstruction}
We begin by assessing the reconstruction performance of the Mo3ENet autoencoder under two conditions: (1) no hydrogens in input or output, and calculated on heavy atoms only, and (2) hydrogens in input and output, with mean distances calculated on all atoms.
Some molecular datasets are published with absent, fuzzy, or incomplete hydrogen information. 
It is therefore useful to have embedding models appropriate for either the presence or absence of hydrogens.

\begin{figure*}
\centering
    \includegraphics[width=\textwidth]{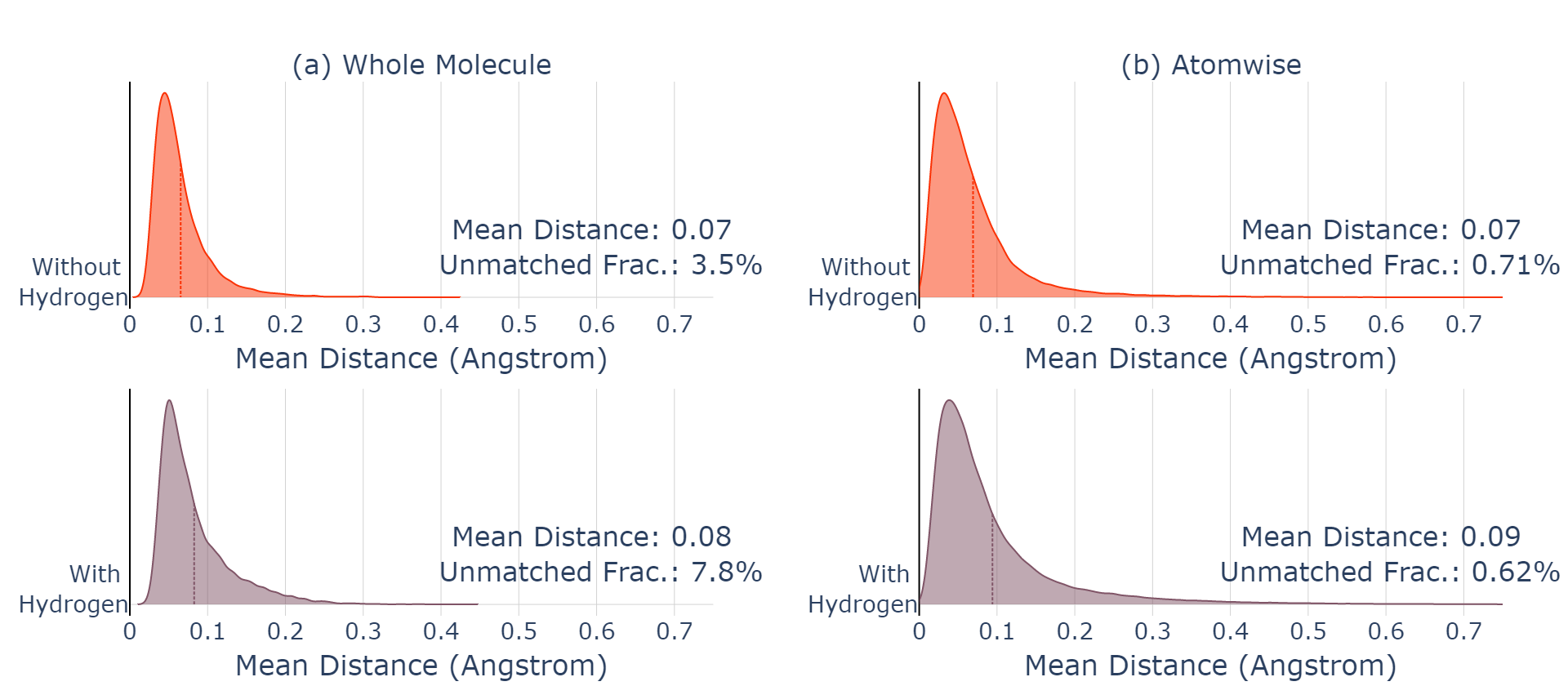}
    \caption{\label{fig:RMSD} Mean distance distribution for Mo3ENet models on the test QM9 dataset (10k samples).
    The models omit or include hydrogen atoms, respectively. 
    In panel (a) we show the mean deviations averaged over the whole molecule, for molecules where every atom was matched by our clustering procedure. We also give the overall average and matched molecule fractions.
    In panel (b) we show the same statistics on a per-atom basis.
}
\end{figure*}

The reconstruction performance is summarized in Figure \ref{fig:RMSD}, where we show the distribution of atom-wise and molecule-averaged mean distances on the test dataset, for the samples where input and output atoms could be uniquely matched by our scaffolding algorithm.
Both models converge well with molecular mean distances $<0.1$\r{A}, and with a reasonably tight distribution about the mean.
We note for each model the mean deviation and fraction of the dataset for which the scaffolding failed to uniquely match inputs to outputs.
These generally correspond to cases where either the output swarm was too diffuse to be neatly clustered, or where one or more atoms were not properly captured, even if the rest of the molecule is correct.
This interpretation is supported by the fact that the atom-by-atom matching fraction is significantly better than that for whole-graphs, indicating that molecules which fail to match are mostly complete, with a relatively small number of imperfectly captured atoms.

A significant positive correlate of error is the fraction of fluorine in each molecule, which should, perhaps, not be surprising since fluorine is by far the rarest atom type in the dataset.
A wider range of molecular backbones would therefore likely boost overall generalization.
Indeed, our experiments indicate longer training runs with more molecules and larger models should yield systematic improvements.

A second and interesting source of significant errors during development were highly symmetric molecules. 
We found that molecules with mirror planes or inversion centres frequently had extremely poor, sometimes practically random, reconstructions. 
This effect arises ironically from one of the design advantages of the model: equivariance. 
The final step in molecule encoding is aggregating atom-wise information to a molecule-scale descriptor, the molecule embedding.
Whether the graph aggregation operation is based on a sum, mean, self-attention, or combination of the above, it requires summing of atom embedding vectors.
In an equivariant model, if two atoms have identical local environments and are located opposite one another by a inversion centre, they will have identical node vectors, up to a global rotation; when summed, they exactly cancel, giving the decoder downstream no information at all to reconstruct the molecule.

Fortunately, we found empirically this could be ameliorated by injecting a tiny amount of noise into the atom positions (<0.01\r{A} on average), lifting the offending symmetry and allowing the encoder to pass nonzero information to the decoder.
Many of the lowest quality reconstructions, even after this patch, are still high-symmetry molecules, indicating it is still more difficult to learn their structures, but their reconstructions are still generally good and mean distances are within the normal range of the data.

\subsection{Embedding analysis}

\begin{figure*}
\centering
    \includegraphics[trim=0 80 0 80, clip, width=1\textwidth]{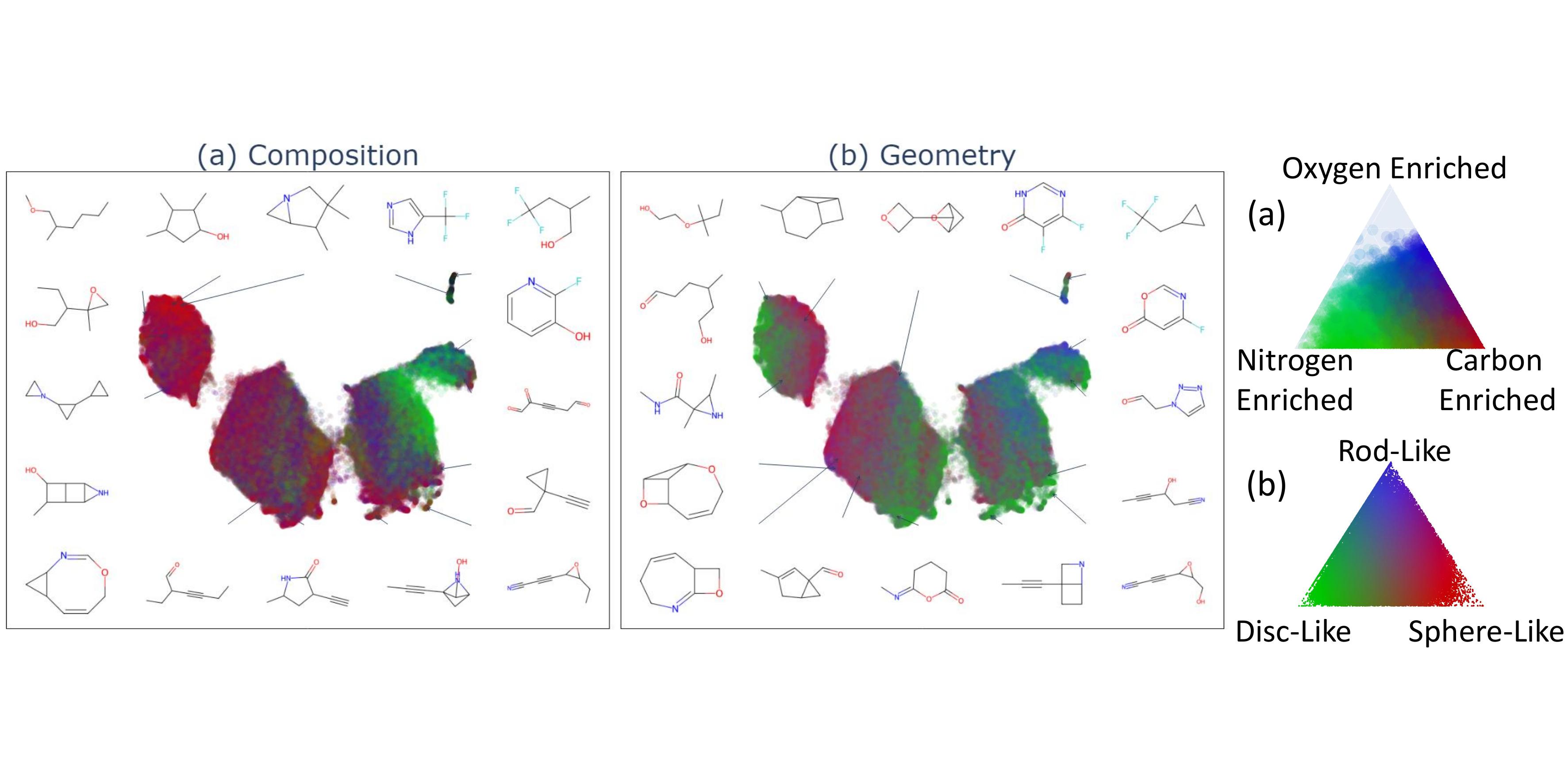}
    \caption{\label{fig:latents} The UMAP decomposition for the full QM9 dataset according to our autoencoder trained with hydrogens, with the x and y axes the UMAP reduced dimensions of the 64 dimensional encoder embedding. 
    Points are color coded according to, (a) compositional (atom fraction) and, (b) geometric molecular (principal inertial ratios) factors, with legends for each on the right hand side.
    }
\end{figure*}

In Figure \ref{fig:latents}, we map the encoder embedding in a reduced dimensional space.
There is an open choice of which form of the embedding to visualize here, the full vector embedding, the scalarized embedding which includes a learned angular projection, or simply the Cartesian norms of the vector embedding.
We chose the scalarized output (details of `scalarizer' in the appendix), as it should contain richer information than the vector norms, including primarily the relative directions of vector components, and is invariant to molecule pose, which could over-complicate the mapping.

We employ UMAP~\cite{mcinnes2018umap}, a dimensionality reduction algorithm, similar in spirit to principal component analysis~\cite{pearson1901liii}, or t-distributed stochastic neighborhood embedding~\cite{hinton2002stochastic}, which attempts to preserve relative distances between samples, such that samples that are near or far in the high dimensional latent space are similarly placed in the reduced dimension.
For visualization purposes, we generated 2-component projections  from the scalarized embedding vectors of the full QM9 dataset, shown in Figure~\ref{fig:latents}(a)-(b), with colors assigned according to (a) relative atomic compositions and (b) shapes.
In the margins, we visualize molecules connected by arrows to their respective points in the embedding space, to give a qualitative sense of the types of molecules clustered in each region.

In Figure \ref{fig:latents}(a) we see quite distinctive clustering following relative elemental enrichment, particularly of nitrogenous compounds. 
In Figure \ref{fig:latents}(b), we see that the relative molecular geometries, as expressed by principal moments of inertia are also differentiated, indicating separation according to molecular shapes.
Overall, this analysis shows that the model is learning a chemically meaningful embedding, which incorporates information on molecular composition and structure, as opposed to simply memorizing unconnected molecules in a large latent space.

\subsection{Property prediction}
\begin{figure*}
\centering
    \includegraphics[trim=540 0 540 0, clip, width=\textwidth]{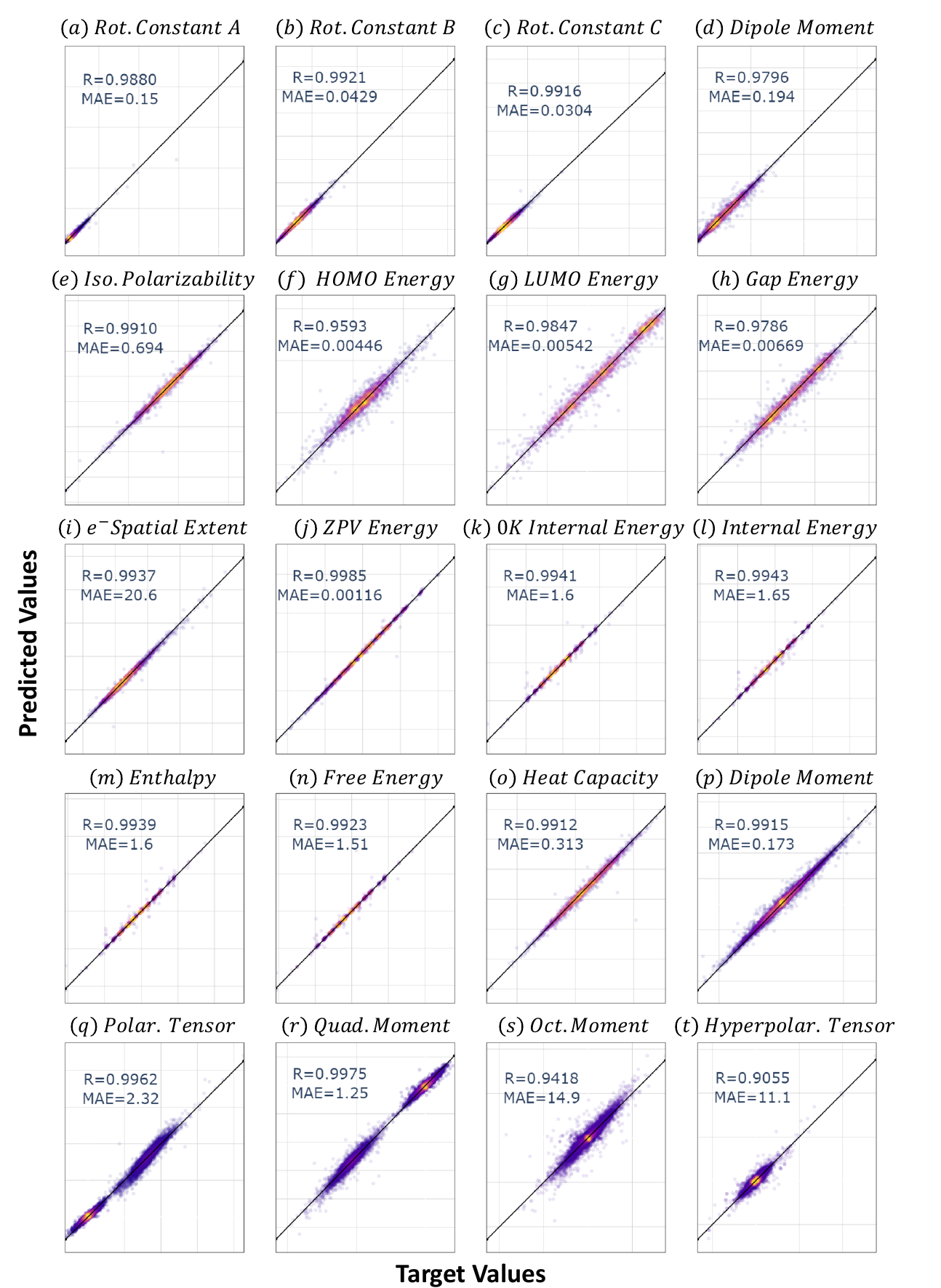}
    \caption{\label{fig:qm9_properties} Predicted vs. true values on the test dataset (20\% of sample) for the 15 QM9 scalar properties, and 5 QM9s vector and tensor properties, with associated correlation coefficients and mean absolute errors.
    Thermodynamic quantities are evaluated at 298K unless otherwise stated.
    Detailed plots for each property are provided in the appendix.
    }
\end{figure*}

We have proven above that our molecule representation is complete and generalizes well to unseen QM9-like conformers.
Given that the molecular embedding generated by the encoder is rich enough to reconstruct the molecule in full, it should serve as a good representation from which to predict molecule properties.
Therefore, to demonstrate further its utility and generalizability, and begin to explore the Mo3ENet's behavior on downstream modelling tasks, we undertake, below, scalar, vector, and tensor prediction tasks on 20 molecular properties.

For the prediction of the 15 scalar properties of the QM9 molecules, we trained separate multilayer perceptrons as regression heads on a frozen pretrained encoder.
For the prediction of the 5 QM9s vector and tensor molecule properties~\cite{zou2023deep}, we trained equivariant multilayer perceptron regression models in the same way. 
QM9s re-optimizes the QM9 molecular geometries, adding dipole and multipole moments, and polaraizabilities.
The dataset was split 80:20 into train:test to assess generalization to unseen molecules.
Regression hyperparameters and training details are given in the appendix.
We undertook a small-scale hyperparameter sweep and selected values which gave good test losses on the molecule gap energy, and reused them for all other properties.

We present the regression results in Figure \ref{fig:qm9_properties}, where we see generally very good fits across the board.
More detailed visualizations are given in the appendix.
HOMO, LUMO and gap energies are typically used for benchmarking on this dataset, and with an average MAE 0.00552, we are within a comparable range to the current benchmark value of 0.00467, set by the extensively pre-trained UniMol~\cite{Zhou2023}.
Vector/tensor property prediction is also qualitatively good, although we do only outperform state-of-the-art purpose-built models such as DetaNet~\cite{zou2023deep} on one task, as shown in Table \ref{table:vec_properties}.
\begin{table}[h!]
\begin{center}
\caption{\label{table:vec_properties} MAE comparison on QM9s properties between Mo3ENet and DetaNet.}
\begin{tabular}{||c | c c c c c||} 
 \hline
 Model & Dipole & Polarizability & Hyperpolarizability & Quadrupole & Octupole  \\ [0.5ex] 
 \hline
 Mo3ENet & 0.173 & 2.32 & 1.25 & 14.9 & 11.1  \\ 
 DetaNet & 0.0089 & 0.137 & 2.04 & 0.38 & 6.29  \\
 \hline
\end{tabular}
\end{center}
\end{table}

For all 20 properties, the target:prediction correlations are high, indicating that the learned embedding is both complete in the sense of containing the relevant molecular information, and well enough regularized that models trained on top of it are able to generalize well to the test data. 

We see in our experiments that dataset size is clearly limiting generalization performance, and we observed significant overfitting in training on all properties. 
This is likely due to combination of the difficulty of the regression tasks, the architecture of the regression head, the complexity of the embedding, and imperfections in the Mo3ENet embeddings.

\subsection{Crystal embedding model}

Going one step further downstream, we introduce here a new learning task, taking advantage of the properties of the Mo3ENet encoding.
Given the inherent relatively low-dimensionality of the parameter space of molecular crystals, which includes cell lengths, cell angles, molecular orientation, internal molecular conformation, molecular center-of-geometry location, and space group, we can combine a molecule embedding with only a few such crystal parameters to create an informationally complete representation, and train models in this space.
We call this representation the crystal embedding,
\begin{equation}
    \mathcal{C}_m=m||\mathcal{C},
\end{equation}
for $m$ a molecule embedding or representation, and $\mathcal{C}$ the crystal parameters corresponding to the unit cell box vectors and and molecule placement in the asymmetric unit (see details in appendix).

As a crystal embedding with a sufficiently rich molecule embedding could be used in principle to model any crystal property.
We selected the crystal intermolecular potential energy for this first task, as it is (1) relatively common and well-understood, and (2) quite physically rich. 
Molecular crystal energies depend in very complex ways on the details of intra- and inter-molecular structure, with, for example, a global rotation of a molecule of only a few degrees potentially resulting in severe interatomic clashes.
Going further, the types of intermolecular interactions a molecule will participate in depend on its composition (presence of relevant functional groups) and conformation.
Highly accurate prediction of crystal energies will therefore only be possible with a molecule embedding incorporating molecular conformation, composition, and pose, such as that from Mo3ENet. 

We trained MLP regression heads to predict two types of qualitatively different intermolecular potentials, and compared the autoencoder embedding to a series of simpler molecule embeddings, to observe trends in model performance with increasing representational detail.

The first intermolecular energy function is a Buckingham potential,
\begin{equation}
    E_{BH} = \sum_{ij}A\cdot e^{-Bd_{ij}} - \frac{C}{d_{ij}^6},
\end{equation}
with the sum taken over intermolecular edges, $d_{ij}$ the distance between atoms $i$ and $j$, divided by the sum of their respective van der Waals radii, and A, B, and C constants (-3.26$E$, $1.0\text{\AA}^{-1}$ , -0.2$E\text{\AA}^{6}$, for $E$ an arbitrary energy unit) selected for the function minimum to qualitatively resemble 12-6 Lennard-Jones potential, with softer intermolecular repulsion.

The second crystal energy for this benchmark is a MACE machine learned interatomic potential (MLIP) for atomistic systems~\cite{batatia2023foundation}, specifically the `medium-sized' MACE-MPA-0 model, targeted at materials. 
Extensively pretrained, MACE-type models have shown reasonable performance on molecular and materials data. 
We use this model here not to compute highly accurate energies for molecular crystals but to capture far more nuanced and complex intermolecular interactions compared to a simple interatomic pair potential.
To calculate the crystal lattice energy within a rigid molecule approximation, we compute separately the MACE-MPA-0 energy for a molecular crystal and gas phase molecule (in the crystal conformation without relaxation), and take the difference,
\begin{equation}
    E_{lattice}= \frac{E_{crystal}}{Z}- E_{gas},
\end{equation}
with $Z$ the number of molecules per unit cell.

We test four molecule embedding schemes, (1) an empty embedding, to challenge what the model can do with only the crystal lattice information (2) the molecule volume, to give the model some insight into crystal densities, (3) principal inertial vectors (eigenvectors and moments of the inertial tensor, equivalent to an ellipsoid description), which has been shown to be a good descriptor of crystal planes~\cite{Galanakis2024}, and (4) Mo3ENet encoding.
These embeddings are combined with the 12 crystal parameters for a rigid, $Z'=1$ molecular crystal, $\mathcal{C}$ (details in the appendix), to form a representation for a crystal.

A dataset of $\sim$2.4M crystals in the $P$1 space group, comprising $\sim$220k unique molecules was synthesized and embedded according to these four schemes, and used to train an MLP (hyperparameters in the appendix) to predict the crystal energy.
An additional 63k structures were synthesized for validation.

Crystal sample generation comprises the following steps: (1) generate a random conformer, (2) sample random crystal parameters and build a supercell of the relevant crystal (see SM for details), (3) optimize the crystal parameters against a Lennard-Jones energy function, holding the intramolecular degrees of freedom fixed, and add the final structure to the dataset. For additional diversity, we add 10 non-optimal samples per optimized crystal, each generated by adding a small amount of Gaussian noise to the optimized cell parameters.
The resulting dataset covers a wide range of chemical structures, molecular conformations, and crystal topologies.

\begin{figure*}
\centering
    \includegraphics[trim=0 65 0 65, clip,width=\textwidth]{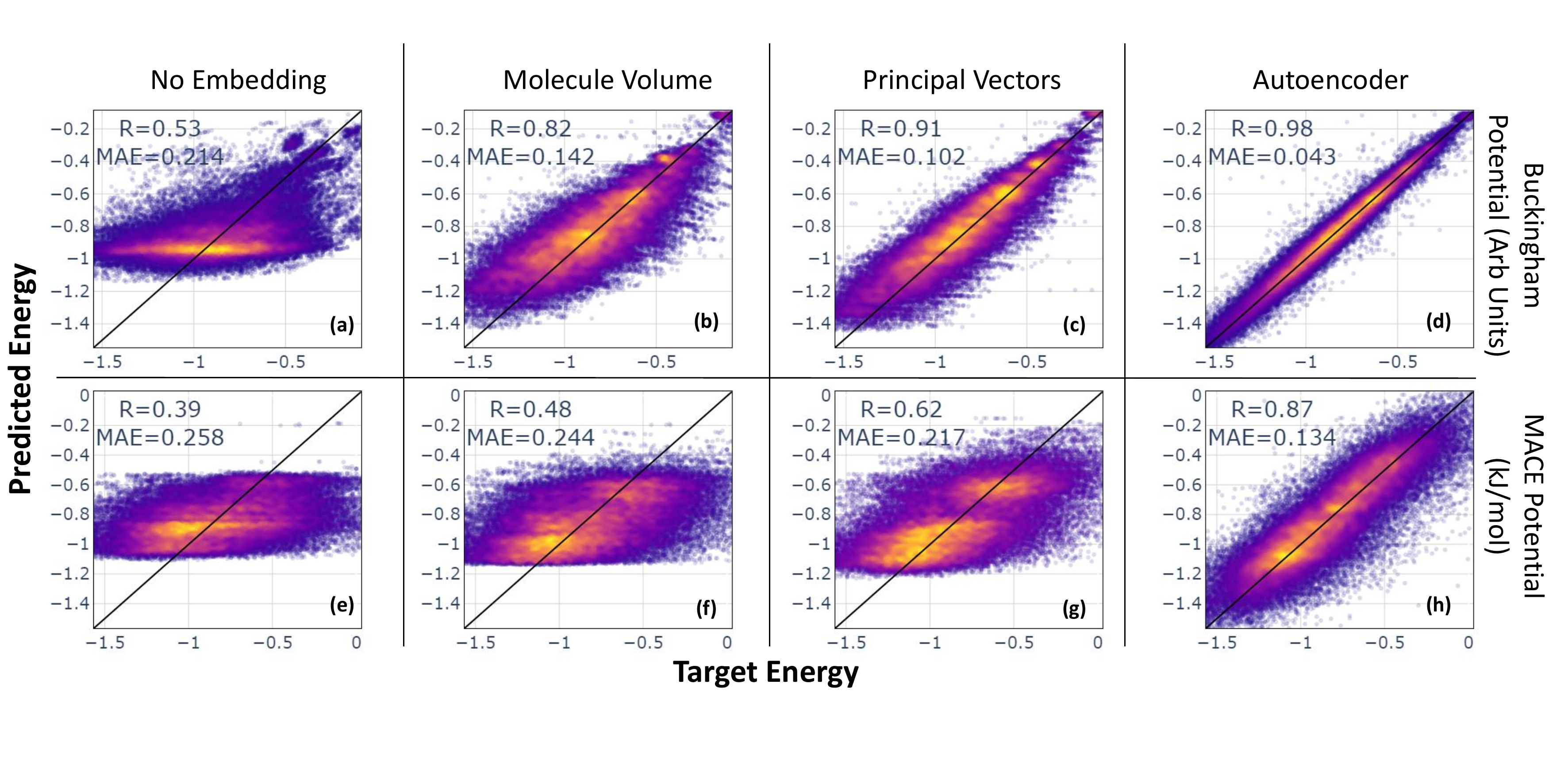}
    \caption{\label{fig:crystal_loss} Parity plots for the crystal embedding energy regression task on the validation set, for several molecule embeddings and two energy functions.
    Subplots (a-d) show the regression results on a Buckingham potential, and (e-h) the same on the MACE MPA-0 potential.
    }
\end{figure*}

We observe in Figure~\ref{fig:crystal_loss} sensible trends in the crystal embedding regression task.
The relatively basic Buckingham energies are possible to estimate, at least somewhat accurately, given only knowledge of molecule size and unit cell volume, assuming minimal intermolecular atomic clashes, a reasonable assumption for this dataset, given its construction procedure.
This is evidenced in subplot (b), where we already see that merely given the molecule volume, a crystal embedding model achieves R=0.76.
In (c), and (d), as we increase the richness of the molecule representation, we see also the regression performance improves, as it is given access to the molecule shape and global orientation (principal vectors), and finally all the atom positions (autoencoder).

For the prediction of MACE MPA-0 lattice energies, we see the detailed information in the autoencoder embedding is much more important for accurate energy prediction.
While the Buckingham potential is not very sensitive to the details of individual atom types and positions, the MACE MPA-0 potential, trained on density functional theory calculations, should be much more sensitive to the intricacies of particular atoms and environments. 
The principal vector embedding outperforms again the volume only or empty embeddings, but none give respectable regression results for the lattice energies. 
Only the autoencoder embedding with its detailed information on molecular position and pose, including the absolute spatial coordinates of all the atoms, is rich enough to achieve a reasonable accuracy.

In terms of training performance,  we observed rapid saturation of the loss and minimal overfitting of the non-autoencoder embeddings due to the low resolution of any molecular features.
The crystal embedding model trained on the autoencoder embedding, on the other hand, continues to improve with training time and increases in dataset size, and so we project its peak performance is even better than what we show in Figure~\ref{fig:crystal_loss}.
We are satisfied the given results make the desired point, and so we chose not to synthesize larger and larger datasets.

\section{\label{sec:Discussion}Discussion}

We introduce and evaluate a new autoencoder architecture and reconstruction loss for the unsupervised learning of multi-type point clouds. 
This tool `solves' the molecule representation problem in the limit of large data, in the sense that it can encode type, conformation, and pose information of a molecule in a provably complete embedding, via a sufficiently large and well trained autoencoder model.
The practical limitations of our model are training time and dataset composition.
To train Mo3ENet on a new class or distribution of molecules requires only sourcing or synthesizing a large and diverse pool of the relevant structures.
The embedding quality on any unseen samples can be trivially evaluated by decoding the sample and computing the reconstruction loss.

Beyond high-fidelity reconstructions, we further proved the quality of the molecule embedding as an input for regressing the twenty QM9/QM9s molecule properties to high accuracy.
We also developed a new type of crystal model, combining molecule embeddings with low-dimensional crystal parameter embeddings to predict properties of molecular crystal structures.
We compared our autoencoder to simpler molecular embeddings and found the increased richness of its molecule representation, including atom types, molecular conformation and pose, was crucial for making accurate crystal energy predictions. Furthermore, the embedding is generally useful for downstream tasks where single molecules need to be represented in the context of larger material structures.

The autoencoder's performance on molecule reconstruction, molecule property regression, and crystal modelling, combined with our qualitative embedding analyses, where we show visually that the model is clustering functional groups and structural motifs in its embedding space, indicate the model is not merely memorizing the input molecules, but generalizing across the space of molecular structures.

This work opens the door to several interesting applications. 
Such a pretrained molecule embedding will be valuable in future exploration of molecule clusters and materials.
For example, in the context of molecular crystal structure prediction, this tool allows us to losslessly coarse-grain molecules to vector embeddings, while retaining the ability to rotate or invert the representations.
This is potentially quite useful in the modelling of molecular clusters or crystals, as we have seen.
This embedding is also potentially the linchpin in the inverse learning problem, i.e., the search, via a generative model, for low energy crystal parameters $\mathcal{C}$, for a given molecular conformer.
Via training on larger and more diverse datasets, we can develop a universal/foundational representation model for high-fidelity encoding of molecule and pose information. 
Such a tool could be useful for molecule property prediction, similarity analysis/clustering, and conditioning for downstream learning tasks.
The speed and generality of our approach also make it attractive as a new type of generative model.
Finally, The freedom of the decoder to generate structures with arbitrary numbers, types, and positions of atoms, and the latent space's amenability to conditioning, is tantalizing to explore further.

\section*{Acknowledgements}
The authors would like to thank Mihail Bogojeski for valuable discussions.
JR acknowledges financial support from the Deutsche Forschungsgemeinschaft (DFG) through the Heisenberg Programme project 428315600. MK, JR, and MET acknowledge funding from grants from the National Science Foundation, DMR-2118890, and MET from CHE-1955381. This work was supported in part through the NYU IT High Performance Computing resources, services, and staff expertise. The Flatiron Institute is a division of the Simons Foundation.

\bibliography{main_v3}

\begin{thebibliography}{72}
\providecommand{\natexlab}[1]{#1}
\providecommand{\url}[1]{\texttt{#1}}
\expandafter\ifx\csname urlstyle\endcsname\relax
  \providecommand{\doi}[1]{doi: #1}\else
  \providecommand{\doi}{doi: \begingroup \urlstyle{rm}\Url}\fi

\bibitem[{\v{Z}}ugec et~al.(2024){\v{Z}}ugec, Geilhufe, and Lon{\v{c}}ari{\'c}]{zugec2024}
Ivan {\v{Z}}ugec, R~Matthias Geilhufe, and Ivor Lon{\v{c}}ari{\'c}.
\newblock Global machine learning potentials for molecular crystals.
\newblock \emph{The Journal of chemical physics}, 160\penalty0 (15), 2024.

\bibitem[Kilgour et~al.(2023)Kilgour, Rogal, and Tuckerman]{kilgour2023}
Michael Kilgour, Jutta Rogal, and Mark Tuckerman.
\newblock Geometric deep learning for molecular crystal structure prediction.
\newblock \emph{Journal of chemical theory and computation}, 19\penalty0 (14):\penalty0 4743--4756, 2023.

\bibitem[Carpenter and Grunwald(2021)]{Carpenter2021}
John~E Carpenter and Michael Grunwald.
\newblock Pre-nucleation clusters predict crystal structures in models of chiral molecules.
\newblock \emph{Journal of the American Chemical Society}, 143\penalty0 (51):\penalty0 21580--21593, 2021.

\bibitem[Egorova et~al.(2020)Egorova, Hafizi, Woods, and Day]{Egorova2020}
Olga Egorova, Roohollah Hafizi, David~C. Woods, and Graeme~M. Day.
\newblock {Multifidelity Statistical Machine Learning for Molecular Crystal Structure Prediction}.
\newblock \emph{Journal of Physical Chemistry A}, 124\penalty0 (39):\penalty0 8065--8078, 2020.

\bibitem[Cersonsky et~al.(2023)Cersonsky, Pakhnova, Engel, and Ceriotti]{Cersonsky2023}
Rose~K. Cersonsky, Maria Pakhnova, Edgar~A. Engel, and Michele Ceriotti.
\newblock {A data-driven interpretation of the stability of organic molecular crystals}.
\newblock \emph{Chemical Science}, 14\penalty0 (5):\penalty0 1272--1285, 2023.

\bibitem[Tom et~al.(2020)Tom, Rose, Bier, O'Brien, V{\'{a}}zquez-Mayagoitia, and Marom]{Tom2020}
Rithwik Tom, Timothy Rose, Imanuel Bier, Harriet O'Brien, {\'{A}}lvaro V{\'{a}}zquez-Mayagoitia, and Noa Marom.
\newblock {Genarris 2.0: A random structure generator for molecular crystals}.
\newblock \emph{Computer Physics Communications}, 250:\penalty0 107170, 2020.

\bibitem[Defever et~al.(2019)Defever, Targonski, Hall, Smith, and Sarupria]{Defever2019}
Ryan~S. Defever, Colin Targonski, Steven~W. Hall, Melissa~C. Smith, and Sapna Sarupria.
\newblock {A generalized deep learning approach for local structure identification in molecular simulations}.
\newblock \emph{Chemical Science}, 10\penalty0 (32):\penalty0 7503--7515, 2019.

\bibitem[Kadan et~al.(2023)Kadan, Ryczko, Wildman, Wang, Roitberg, and Yamazaki]{Kadan2023}
Amit Kadan, Kevin Ryczko, Andrew Wildman, Rodrigo Wang, Adrian Roitberg, and Takeshi Yamazaki.
\newblock Accelerated organic crystal structure prediction with genetic algorithms and machine learning.
\newblock \emph{Journal of Chemical Theory and Computation}, 19\penalty0 (24):\penalty0 9388--9402, 2023.

\bibitem[K{\"{o}}hler et~al.(2023)K{\"{o}}hler, Invernizzi, de~Haan, and No{\'{e}}]{Kohler2023}
Jonas K{\"{o}}hler, Michele Invernizzi, Pim de~Haan, and Frank No{\'{e}}.
\newblock {Rigid Body Flows for Sampling Molecular Crystal Structures}.
\newblock \emph{Proceedings of Machine Learning Research}, 202:\penalty0 17301--17326, 2023.

\bibitem[Apostolakis et~al.(2001)Apostolakis, Hofmann, and Lengauer]{Apostolakis2001}
Joannis Apostolakis, Detlef Walter~Maria Hofmann, and Thomas Lengauer.
\newblock {Derivation of a scoring function for crystal structure prediction}.
\newblock \emph{Acta Crystallographica Section A: Foundations of Crystallography}, 57\penalty0 (4):\penalty0 442--450, 2001.

\bibitem[Guo et~al.(2022)Guo, Guo, Nan, Tian, Iyer, Ma, Wiest, Zhang, Wang, Zhang, et~al.]{guo2022graph}
Zhichun Guo, Kehan Guo, Bozhao Nan, Yijun Tian, Roshni~G Iyer, Yihong Ma, Olaf Wiest, Xiangliang Zhang, Wei Wang, Chuxu Zhang, et~al.
\newblock Graph-based molecular representation learning.
\newblock \emph{arXiv preprint arXiv:2207.04869}, 2022.

\bibitem[Wieder et~al.(2020)Wieder, Kohlbacher, Kuenemann, Garon, Ducrot, Seidel, and Langer]{wieder2020compact}
Oliver Wieder, Stefan Kohlbacher, M{\'e}laine Kuenemann, Arthur Garon, Pierre Ducrot, Thomas Seidel, and Thierry Langer.
\newblock A compact review of molecular property prediction with graph neural networks.
\newblock \emph{Drug Discovery Today: Technologies}, 37:\penalty0 1--12, 2020.

\bibitem[Zhou et~al.(2022)Zhou, Gao, Ding, Zheng, Xu, Wei, Zhang, and Ke]{Zhou2023}
Gengmo Zhou, Zhifeng Gao, Qiankun Ding, Hang Zheng, Hongteng Xu, Zhewei Wei, Linfeng Zhang, and Guolin Ke.
\newblock Uni-mol: A universal 3d molecular representation learning framework.
\newblock In \emph{International Conference on Learning Representations}, 2022.

\bibitem[Fang et~al.(2022)Fang, Liu, Lei, He, Zhang, Zhou, Wang, Wu, and Wang]{fang2022geometry}
Xiaomin Fang, Lihang Liu, Jieqiong Lei, Donglong He, Shanzhuo Zhang, Jingbo Zhou, Fan Wang, Hua Wu, and Haifeng Wang.
\newblock Geometry-enhanced molecular representation learning for property prediction.
\newblock \emph{Nature Machine Intelligence}, 4\penalty0 (2):\penalty0 127--134, 2022.

\bibitem[Zhu et~al.(2022)Zhu, Xia, Wu, Xie, Qin, Zhou, Li, and Liu]{zhu2022unified}
Jinhua Zhu, Yingce Xia, Lijun Wu, Shufang Xie, Tao Qin, Wengang Zhou, Houqiang Li, and Tie-Yan Liu.
\newblock Unified 2d and 3d pre-training of molecular representations.
\newblock In \emph{Proceedings of the 28th ACM SIGKDD conference on knowledge discovery and data mining}, pages 2626--2636, 2022.

\bibitem[Liu et~al.(2021)Liu, Wang, Liu, Lasenby, Guo, and Tang]{liu2021pre}
Shengchao Liu, Hanchen Wang, Weiyang Liu, Joan Lasenby, Hongyu Guo, and Jian Tang.
\newblock Pre-training molecular graph representation with 3d geometry.
\newblock \emph{arXiv preprint arXiv:2110.07728}, 2021.

\bibitem[Ni et~al.(2024)Ni, Feng, Hong, Sun, Ma, Ma, Ye, and Lan]{ni2024pre}
Yuyan Ni, Shikun Feng, Xin Hong, Yuancheng Sun, Wei-Ying Ma, Zhi-Ming Ma, Qiwei Ye, and Yanyan Lan.
\newblock Pre-training with fractional denoising to enhance molecular property prediction.
\newblock \emph{Nature Machine Intelligence}, 6\penalty0 (10):\penalty0 1169--1178, 2024.

\bibitem[Zaidi et~al.(2022)Zaidi, Schaarschmidt, Martens, Kim, Teh, Sanchez-Gonzalez, Battaglia, Pascanu, and Godwin]{zaidi2022pre}
Sheheryar Zaidi, Michael Schaarschmidt, James Martens, Hyunjik Kim, Yee~Whye Teh, Alvaro Sanchez-Gonzalez, Peter Battaglia, Razvan Pascanu, and Jonathan Godwin.
\newblock Pre-training via denoising for molecular property prediction.
\newblock \emph{arXiv preprint arXiv:2206.00133}, 2022.

\bibitem[Zhang et~al.(2023)Zhang, Fan, Liu, Huang, Zhao, Huang, and Liu]{zhang2023expressive}
Bingxu Zhang, Changjun Fan, Shixuan Liu, Kuihua Huang, Xiang Zhao, Jincai Huang, and Zhong Liu.
\newblock The expressive power of graph neural networks: A survey.
\newblock \emph{arXiv preprint arXiv:2308.08235}, 2023.

\bibitem[Corso et~al.(2024)Corso, Stark, Jegelka, Jaakkola, and Barzilay]{corso2024graph}
Gabriele Corso, Hannes Stark, Stefanie Jegelka, Tommi Jaakkola, and Regina Barzilay.
\newblock Graph neural networks.
\newblock \emph{Nature Reviews Methods Primers}, 4\penalty0 (1):\penalty0 17, 2024.

\bibitem[Sch{\"u}tt et~al.(2021)Sch{\"u}tt, Unke, and Gastegger]{Schutt2021}
Kristof Sch{\"u}tt, Oliver Unke, and Michael Gastegger.
\newblock Equivariant message passing for the prediction of tensorial properties and molecular spectra.
\newblock In \emph{International Conference on Machine Learning}, pages 9377--9388, 2021.

\bibitem[Le et~al.(2022)Le, Noe, and Clevert]{Le2022}
Tuan Le, Frank Noe, and Djork-Arn{\'e} Clevert.
\newblock Representation learning on biomolecular structures using equivariant graph attention.
\newblock In \emph{Learning on Graphs Conference}, pages 1--17, 2022.

\bibitem[Deng et~al.(2021)Deng, Litany, Duan, Poulenard, Tagliasacchi, and Guibas]{Deng2022}
Congyue Deng, Or~Litany, Yueqi Duan, Adrien Poulenard, Andrea Tagliasacchi, and Leonidas~J Guibas.
\newblock Vector neurons: A general framework for so (3)-equivariant networks.
\newblock In \emph{Proceedings of the IEEE/CVF International Conference on Computer Vision}, pages 12200--12209, 2021.

\bibitem[Pozdnyakov and Ceriotti(2022)]{Pozdnyakov2022}
Sergey~N. Pozdnyakov and Michele Ceriotti.
\newblock {Incompleteness of graph neural networks for points clouds in three dimensions}.
\newblock \emph{Machine Learning: Science and Technology}, 3\penalty0 (4):\penalty0 1--7, 2022.

\bibitem[Jing et~al.(2022)Jing, Corso, Chang, Barzilay, and Jaakkola]{Jing2022}
Bowen Jing, Gabriele Corso, Jeffrey Chang, Regina Barzilay, and Tommi Jaakkola.
\newblock {Torsional Diffusion for Molecular Conformer Generation}.
\newblock \emph{Advances in Neural Information Processing Systems}, 35:\penalty0 1--5, 2022.

\bibitem[Xu et~al.(2021)Xu, Yu, Song, Shi, Ermon, and Tang]{xu2022geodiff}
Minkai Xu, Lantao Yu, Yang Song, Chence Shi, Stefano Ermon, and Jian Tang.
\newblock Geodiff: A geometric diffusion model for molecular conformation generation.
\newblock In \emph{International Conference on Learning Representations}, 2021.

\bibitem[Xu et~al.(2023)Xu, Powers, Dror, Ermon, and Leskovec]{xu2023geometric}
Minkai Xu, Alexander~S Powers, Ron~O Dror, Stefano Ermon, and Jure Leskovec.
\newblock Geometric latent diffusion models for 3d molecule generation.
\newblock In \emph{International Conference on Machine Learning}, pages 38592--38610, 2023.

\bibitem[Hoogeboom et~al.(2022)Hoogeboom, Satorras, Vignac, and Welling]{hoogeboom2022equivariant}
Emiel Hoogeboom, V{\i}ctor~Garcia Satorras, Cl{\'e}ment Vignac, and Max Welling.
\newblock Equivariant diffusion for molecule generation in 3d.
\newblock In \emph{International conference on machine learning}, pages 8867--8887, 2022.

\bibitem[Ruddigkeit et~al.(2012)Ruddigkeit, Van~Deursen, Blum, and Reymond]{ruddigkeit2012enumeration}
Lars Ruddigkeit, Ruud Van~Deursen, Lorenz~C Blum, and Jean-Louis Reymond.
\newblock Enumeration of 166 billion organic small molecules in the chemical universe database gdb-17.
\newblock \emph{Journal of chemical information and modeling}, 52\penalty0 (11):\penalty0 2864--2875, 2012.

\bibitem[Ramakrishnan et~al.(2014)Ramakrishnan, Dral, Rupp, and Von~Lilienfeld]{ramakrishnan2014quantum}
Raghunathan Ramakrishnan, Pavlo~O Dral, Matthias Rupp, and O~Anatole Von~Lilienfeld.
\newblock Quantum chemistry structures and properties of 134 kilo molecules.
\newblock \emph{Scientific data}, 1\penalty0 (1):\penalty0 1--7, 2014.

\bibitem[Tingle et~al.(2023)Tingle, Tang, Castanon, Gutierrez, Khurelbaatar, Dandarchuluun, Moroz, and Irwin]{tingle2023zinc}
Benjamin~I Tingle, Khanh~G Tang, Mar Castanon, John~J Gutierrez, Munkhzul Khurelbaatar, Chinzorig Dandarchuluun, Yurii~S Moroz, and John~J Irwin.
\newblock Zinc-22- a free multi-billion-scale database of tangible compounds for ligand discovery.
\newblock \emph{Journal of chemical information and modeling}, 63\penalty0 (4):\penalty0 1166--1176, 2023.

\bibitem[Qi et~al.(2017)Qi, Su, Mo, and Guibas]{Qi2017}
Charles~R Qi, Hao Su, Kaichun Mo, and Leonidas~J Guibas.
\newblock Pointnet: Deep learning on point sets for 3d classification and segmentation.
\newblock In \emph{Proceedings of the IEEE conference on computer vision and pattern recognition}, pages 652--660, 2017.

\bibitem[Tchapmi et~al.(2019)Tchapmi, Kosaraju, Rezatofighi, Reid, and Savarese]{Tchapmi2019}
Lyne~P Tchapmi, Vineet Kosaraju, Hamid Rezatofighi, Ian Reid, and Silvio Savarese.
\newblock Topnet: Structural point cloud decoder.
\newblock In \emph{Proceedings of the IEEE/CVF conference on computer vision and pattern recognition}, pages 383--392, 2019.

\bibitem[Yang et~al.(2018)Yang, Feng, Shen, and Tian]{Yang2018}
Yaoqing Yang, Chen Feng, Yiru Shen, and Dong Tian.
\newblock Foldingnet: Point cloud auto-encoder via deep grid deformation.
\newblock In \emph{Proceedings of the IEEE conference on computer vision and pattern recognition}, pages 206--215, 2018.

\bibitem[Guo et~al.(2020)Guo, Wang, Hu, Liu, Liu, and Bennamoun]{guo2020deep}
Yulan Guo, Hanyun Wang, Qingyong Hu, Hao Liu, Li~Liu, and Mohammed Bennamoun.
\newblock Deep learning for 3d point clouds: A survey.
\newblock \emph{IEEE transactions on pattern analysis and machine intelligence}, 43\penalty0 (12):\penalty0 4338--4364, 2020.

\bibitem[Zhang et~al.(2019)Zhang, Zhao, Chen, and Lu]{zhang2019review}
Jiaying Zhang, Xiaoli Zhao, Zheng Chen, and Zhejun Lu.
\newblock A review of deep learning-based semantic segmentation for point cloud.
\newblock \emph{IEEE access}, 7:\penalty0 179118--179133, 2019.

\bibitem[Suchde et~al.(2023)Suchde, Jacquemin, and Davydov]{suchde2023point}
Pratik Suchde, Thibault Jacquemin, and Oleg Davydov.
\newblock Point cloud generation for meshfree methods: An overview.
\newblock \emph{Archives of Computational Methods in Engineering}, 30\penalty0 (2):\penalty0 889--915, 2023.

\bibitem[Li et~al.(2019)Li, Bi, and Lee]{Li2019}
Jiaxin Li, Yingcai Bi, and Gim~Hee Lee.
\newblock Discrete rotation equivariance for point cloud recognition.
\newblock In \emph{International conference on robotics and automation}, pages 7269--7275. IEEE, 2019.

\bibitem[Vignac et~al.(2020)Vignac, Loukas, and Frossard]{Vignac2020}
Clement Vignac, Andreas Loukas, and Pascal Frossard.
\newblock Building powerful and equivariant graph neural networks with structural message-passing.
\newblock \emph{Advances in neural information processing systems}, 33:\penalty0 14143--14155, 2020.

\bibitem[Keriven and Peyr{\'e}(2019)]{Keriven2019}
Nicolas Keriven and Gabriel Peyr{\'e}.
\newblock Universal invariant and equivariant graph neural networks.
\newblock \emph{Advances in Neural Information Processing Systems}, 32, 2019.

\bibitem[Finzi et~al.(2021)Finzi, Welling, and Wilson]{finzi2021practical}
Marc Finzi, Max Welling, and Andrew~Gordon Wilson.
\newblock A practical method for constructing equivariant multilayer perceptrons for arbitrary matrix groups.
\newblock In \emph{International conference on machine learning}, pages 3318--3328. PMLR, 2021.

\bibitem[Liao and Smidt(2022)]{liao2022equiformer}
Yi-Lun Liao and Tess Smidt.
\newblock Equiformer: Equivariant graph attention transformer for 3d atomistic graphs.
\newblock \emph{arXiv preprint arXiv:2206.11990}, 2022.

\bibitem[Satorras et~al.(2021)Satorras, Hoogeboom, and Welling]{satorras2021n}
V{\i}ctor~Garcia Satorras, Emiel Hoogeboom, and Max Welling.
\newblock E (n) equivariant graph neural networks.
\newblock In \emph{International conference on machine learning}, pages 9323--9332, 2021.

\bibitem[Choe et~al.(2021)Choe, Joung, Rameau, Park, and Kweon]{Choe2022}
Jaesung Choe, ByeongIn Joung, Francois Rameau, Jaesik Park, and In~So Kweon.
\newblock Deep point cloud reconstruction.
\newblock In \emph{International Conference on Learning Representations}, 2021.

\bibitem[Zamorski et~al.(2020)Zamorski, Zi{\c{e}}ba, Klukowski, Nowak, Kurach, Stokowiec, and Trzci{\'{n}}ski]{Zamorski2020}
Maciej Zamorski, Maciej Zi{\c{e}}ba, Piotr Klukowski, Rafa{\l} Nowak, Karol Kurach, Wojciech Stokowiec, and Tomasz Trzci{\'{n}}ski.
\newblock {Adversarial autoencoders for compact representations of 3D point clouds}.
\newblock \emph{Computer Vision and Image Understanding}, 193, 2020.

\bibitem[Jiang et~al.(2018)Jiang, Shi, Qi, and Jia]{Jiang2018}
Li~Jiang, Shaoshuai Shi, Xiaojuan Qi, and Jiaya Jia.
\newblock Gal: Geometric adversarial loss for single-view 3d-object reconstruction.
\newblock In \emph{Proceedings of the European conference on computer vision}, pages 802--816, 2018.

\bibitem[Ahmadi et~al.(2024)Ahmadi, Ghanavati, and Rohani]{ahmadi2024machine}
Soroush Ahmadi, Mohammad~Amin Ghanavati, and Sohrab Rohani.
\newblock Machine learning-guided prediction of cocrystals using point cloud-based molecular representation.
\newblock \emph{Chemistry of Materials}, 2024.

\bibitem[Hansen et~al.(2015)Hansen, Biegler, Ramakrishnan, Pronobis, Von~Lilienfeld, Muller, and Tkatchenko]{hansen2015machine}
Katja Hansen, Franziska Biegler, Raghunathan Ramakrishnan, Wiktor Pronobis, O~Anatole Von~Lilienfeld, Klaus-Robert Muller, and Alexandre Tkatchenko.
\newblock Machine learning predictions of molecular properties: Accurate many-body potentials and nonlocality in chemical space.
\newblock \emph{The journal of physical chemistry letters}, 6\penalty0 (12):\penalty0 2326--2331, 2015.

\bibitem[Rupp et~al.(2012)Rupp, Tkatchenko, M{\"u}ller, and Von~Lilienfeld]{rupp2012fast}
Matthias Rupp, Alexandre Tkatchenko, Klaus-Robert M{\"u}ller, and O~Anatole Von~Lilienfeld.
\newblock Fast and accurate modeling of molecular atomization energies with machine learning.
\newblock \emph{Physical review letters}, 108\penalty0 (5):\penalty0 058301, 2012.

\bibitem[Bart{\'o}k et~al.(2013)Bart{\'o}k, Kondor, and Cs{\'a}nyi]{bartok2013representing}
Albert~P Bart{\'o}k, Risi Kondor, and G{\'a}bor Cs{\'a}nyi.
\newblock On representing chemical environments.
\newblock \emph{Physical Review B}, 87\penalty0 (18):\penalty0 184115, 2013.

\bibitem[Musil et~al.(2021)Musil, Grisafi, Bart{\'o}k, Ortner, Cs{\'a}nyi, and Ceriotti]{musil_PhysicsInspiredStructural_2021}
Felix Musil, Andrea Grisafi, Albert~P. Bart{\'o}k, Christoph Ortner, G{\'a}bor Cs{\'a}nyi, and Michele Ceriotti.
\newblock Physics-{{Inspired Structural Representations}} for {{Molecules}} and {{Materials}}.
\newblock \emph{Chem. Rev.}, 121\penalty0 (16):\penalty0 9759--9815, August 2021.

\bibitem[Yang et~al.(2019)Yang, Swanson, Jin, Coley, Eiden, Gao, Guzman-Perez, Hopper, Kelley, Mathea, Palmer, Settels, Jaakkola, Jensen, and Barzilay]{Yang2019}
Kevin Yang, Kyle Swanson, Wengong Jin, Connor Coley, Philipp Eiden, Hua Gao, Angel Guzman-Perez, Timothy Hopper, Brian Kelley, Miriam Mathea, Andrew Palmer, Volker Settels, Tommi Jaakkola, Klavs Jensen, and Regina Barzilay.
\newblock {Analyzing Learned Molecular Representations for Property Prediction}.
\newblock \emph{Journal of Chemical Information and Modeling}, 59\penalty0 (8):\penalty0 3370--3388, 2019.

\bibitem[Wang et~al.(2022)Wang, Wang, Cao, and Barati~Farimani]{wang2022molecular}
Yuyang Wang, Jianren Wang, Zhonglin Cao, and Amir Barati~Farimani.
\newblock Molecular contrastive learning of representations via graph neural networks.
\newblock \emph{Nature Machine Intelligence}, 4\penalty0 (3):\penalty0 279--287, 2022.

\bibitem[Raghunathan and Priyakumar(2022)]{raghunathan2022molecular}
Shampa Raghunathan and U~Deva Priyakumar.
\newblock Molecular representations for machine learning applications in chemistry.
\newblock \emph{International Journal of Quantum Chemistry}, 122\penalty0 (7):\penalty0 e26870, 2022.

\bibitem[You et~al.(2018)You, Ying, Ren, Hamilton, and Leskovec]{You2018}
Jiaxuan You, Rex Ying, Xiang Ren, William Hamilton, and Jure Leskovec.
\newblock Graphrnn: Generating realistic graphs with deep auto-regressive models.
\newblock In \emph{International conference on machine learning}, pages 5708--5717, 2018.

\bibitem[Winter et~al.(2021)Winter, No{\'e}, and Clevert]{Winter2021}
Robin Winter, Frank No{\'e}, and Djork-Arn{\'e} Clevert.
\newblock Auto-encoding molecular conformations.
\newblock \emph{arXiv preprint arXiv:2101.01618}, 2021.

\bibitem[Joo et~al.(2020)Joo, Kim, Yang, and Park]{Joo2020}
Sunghoon Joo, Min~Soo Kim, Jaeho Yang, and Jeahyun Park.
\newblock {Generative Model for Proposing Drug Candidates Satisfying Anticancer Properties Using a Conditional Variational Autoencoder}.
\newblock \emph{ACS Omega}, 5\penalty0 (30):\penalty0 18642--18650, 2020.

\bibitem[Simonovsky and Komodakis(2018)]{Simonovsky2018}
Martin Simonovsky and Nikos Komodakis.
\newblock Graphvae: Towards generation of small graphs using variational autoencoders.
\newblock In \emph{International Conference on Artificial Neural Networks}, pages 412--422, 2018.

\bibitem[Xu et~al.(2022)Xu, Huang, Xu, Lei, and Chen]{Graph2023}
Mingyuan Xu, Weifeng Huang, Min Xu, Jinping Lei, and Hongming Chen.
\newblock 3d conformational generative models for biological structures using graph information-embedded relative coordinates.
\newblock \emph{Molecules}, 28\penalty0 (1):\penalty0 321, 2022.

\bibitem[Ilnicka and Schneider(2023)]{Ilnicka2023}
Agnieszka Ilnicka and Gisbert Schneider.
\newblock Designing molecules with autoencoder networks.
\newblock \emph{Nature Computational Science}, 3\penalty0 (11):\penalty0 922--933, 2023.

\bibitem[Chennakesavalu et~al.(2023)Chennakesavalu, Toomer, and Rotskoff]{Chennakesavalu2023}
Shriram Chennakesavalu, David~J Toomer, and Grant~M Rotskoff.
\newblock Ensuring thermodynamic consistency with invertible coarse-graining.
\newblock \emph{The Journal of Chemical Physics}, 158\penalty0 (12), 2023.

\bibitem[rdk()]{rdkit}
Rdkit: Open-source cheminformatics.
\newblock URL \url{https://www.rdkit.org}.

\bibitem[Axelrod and Gomez-Bombarelli(2022)]{axelrod2022geom}
Simon Axelrod and Rafael Gomez-Bombarelli.
\newblock Geom, energy-annotated molecular conformations for property prediction and molecular generation.
\newblock \emph{Scientific Data}, 9\penalty0 (1):\penalty0 185, 2022.

\bibitem[Isert et~al.(2022)Isert, Atz, Jim{\'e}nez-Luna, and Schneider]{isert2022qmugs}
Clemens Isert, Kenneth Atz, Jos{\'e} Jim{\'e}nez-Luna, and Gisbert Schneider.
\newblock Qmugs, quantum mechanical properties of drug-like molecules.
\newblock \emph{Scientific Data}, 9\penalty0 (1):\penalty0 273, 2022.

\bibitem[McInnes et~al.(2018)McInnes, Healy, and Melville]{mcinnes2018umap}
Leland McInnes, John Healy, and James Melville.
\newblock Umap: Uniform manifold approximation and projection for dimension reduction.
\newblock \emph{arXiv preprint arXiv:1802.03426}, 2018.

\bibitem[Pearson(1901)]{pearson1901liii}
Karl Pearson.
\newblock Liii. on lines and planes of closest fit to systems of points in space.
\newblock \emph{The London, Edinburgh, and Dublin philosophical magazine and journal of science}, 2\penalty0 (11):\penalty0 559--572, 1901.

\bibitem[Hinton and Roweis(2002)]{hinton2002stochastic}
Geoffrey~E Hinton and Sam Roweis.
\newblock Stochastic neighbor embedding.
\newblock In S.~Becker, S.~Thrun, and K.~Obermayer, editors, \emph{Advances in Neural Information Processing Systems}, volume~15 of \emph{NIPS'02}, page 857–864. MIT Press, 2002.

\bibitem[Zou et~al.(2023)Zou, Zhang, Liang, Wei, Leng, Jiang, Luo, and Hu]{zou2023deep}
Zihan Zou, Yujin Zhang, Lijun Liang, Mingzhi Wei, Jiancai Leng, Jun Jiang, Yi~Luo, and Wei Hu.
\newblock A deep learning model for predicting selected organic molecular spectra.
\newblock \emph{Nature Computational Science}, 3\penalty0 (11):\penalty0 957--964, 2023.

\bibitem[Batatia et~al.(2023)Batatia, Benner, Chiang, Elena, Kovács, Riebesell, Advincula, Asta, Baldwin, Bernstein, Bhowmik, Blau, Cărare, Darby, De, Pia, Deringer, Elijošius, El-Machachi, Fako, Ferrari, Genreith-Schriever, George, Goodall, Grey, Han, Handley, Heenen, Hermansson, Holm, Jaafar, Hofmann, Jakob, Jung, Kapil, Kaplan, Karimitari, Kroupa, Kullgren, Kuner, Kuryla, Liepuoniute, Margraf, Magdău, Michaelides, Moore, Naik, Niblett, Norwood, O'Neill, Ortner, Persson, Reuter, Rosen, Schaaf, Schran, Sivonxay, Stenczel, Svahn, Sutton, van~der Oord, Varga-Umbrich, Vegge, Vondrák, Wang, Witt, Zills, and Csányi]{batatia2023foundation}
Ilyes Batatia, Philipp Benner, Yuan Chiang, Alin~M. Elena, Dávid~P. Kovács, Janosh Riebesell, Xavier~R. Advincula, Mark Asta, William~J. Baldwin, Noam Bernstein, Arghya Bhowmik, Samuel~M. Blau, Vlad Cărare, James~P. Darby, Sandip De, Flaviano~Della Pia, Volker~L. Deringer, Rokas Elijošius, Zakariya El-Machachi, Edvin Fako, Andrea~C. Ferrari, Annalena Genreith-Schriever, Janine George, Rhys E.~A. Goodall, Clare~P. Grey, Shuang Han, Will Handley, Hendrik~H. Heenen, Kersti Hermansson, Christian Holm, Jad Jaafar, Stephan Hofmann, Konstantin~S. Jakob, Hyunwook Jung, Venkat Kapil, Aaron~D. Kaplan, Nima Karimitari, Namu Kroupa, Jolla Kullgren, Matthew~C. Kuner, Domantas Kuryla, Guoda Liepuoniute, Johannes~T. Margraf, Ioan-Bogdan Magdău, Angelos Michaelides, J.~Harry Moore, Aakash~A. Naik, Samuel~P. Niblett, Sam~Walton Norwood, Niamh O'Neill, Christoph Ortner, Kristin~A. Persson, Karsten Reuter, Andrew~S. Rosen, Lars~L. Schaaf, Christoph Schran, Eric Sivonxay, Tamás~K. Stenczel, Viktor Svahn, Christopher
  Sutton, Cas van~der Oord, Eszter Varga-Umbrich, Tejs Vegge, Martin Vondrák, Yangshuai Wang, William~C. Witt, Fabian Zills, and Gábor Csányi.
\newblock A foundation model for atomistic materials chemistry.
\newblock 2023.

\bibitem[Galanakis and Tuckerman(2024)]{Galanakis2024}
Nikolaos Galanakis and Mark~E. Tuckerman.
\newblock Rapid prediction of molecular crystal structures using simple topological and physical descriptors.
\newblock \emph{Nature Communications}, 15\penalty0 (1), November 2024.
\newblock ISSN 2041-1723.
\newblock \doi{10.1038/s41467-024-53596-5}.
\newblock URL \url{http://dx.doi.org/10.1038/s41467-024-53596-5}.

\bibitem[Li et~al.(2020)Li, Xiong, Thabet, and Ghanem]{li2020deepergcn}
Guohao Li, Chenxin Xiong, Ali Thabet, and Bernard Ghanem.
\newblock Deepergcn: All you need to train deeper gcns.
\newblock \emph{arXiv preprint arXiv:2006.07739}, 2020.

\bibitem[Loshchilov and Hutter(2018)]{loshchilov2017decoupled}
Ilya Loshchilov and Frank Hutter.
\newblock Decoupled weight decay regularization.
\newblock In \emph{International Conference on Learning Representations}, 2018.

\end{thebibliography}

\newpage

\appendix

\section{Details of Model Architecture}

\begin{figure*}
\renewcommand{\thefigure}{A1}
\centering
    \includegraphics[trim=250 70 250 70, clip, width=\textwidth]{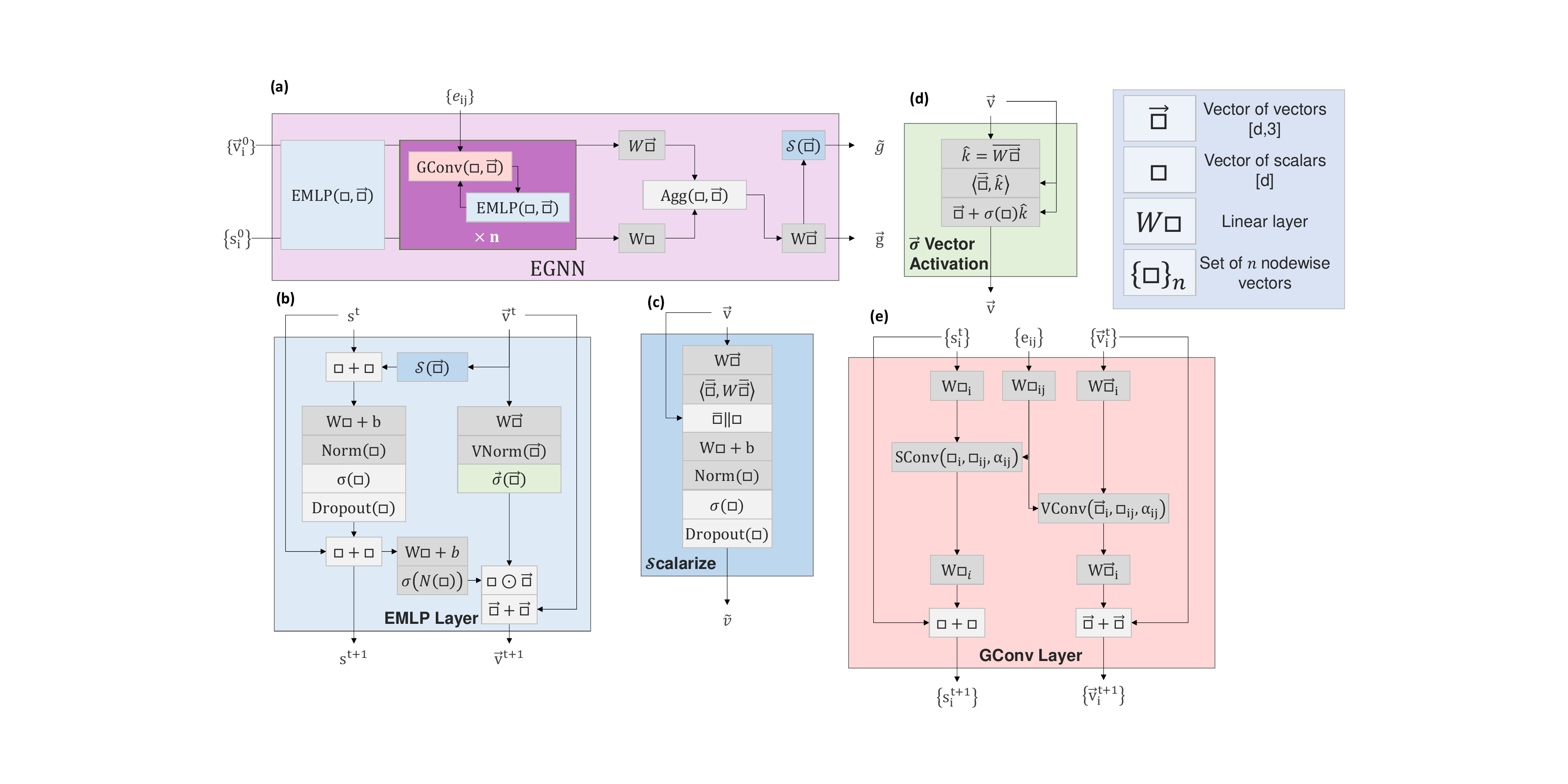}
    \caption{\label{fig:architecture}Panels (a), (b), (c), (d), and (e) outline the architecture for the graph neural net, equivariant multilayer perceptron, scalarizer, vector activation, and graph convolution modules, respectively. 
    Dark grey nodes represent functions with learnable weights, and lighter grey functions without.}
\end{figure*}

The EMLP layer acts on scalar inputs $\bm{s}$ of dimension $[n, k]$, for $n$ the batch index and $k$ the feature depth for scalars, and vector inputs $\bm{\vec{v}}$ of dimension $[n, 3, l]$, for $l$ the feature depth for vectors. 
The scalar track is a standard neural network layer acting on  with linear transformation according to learned weight matrix $W^t$ and bias $b^t$, normalization $\mathrm{N}$, activation $\sigma$, dropout $\mathrm{D}$, and residual connection (addition by $\bm{s}^{t}$), with optional dimension adjustment between layers,
\begin{equation}\label{eq:scalar_forward}
    \bm{s^{t+1}}=\bm{s}^{t} + \mathrm{D}(\sigma(\mathrm{N}(\bm{W}^{t}(\bm{s}^{t}+\tilde{\bm{v}}^{t})+ \bm{b}^{t}))),
\end{equation}
and scalarized representation of the vector track $\bm{\tilde{v}}=\mathcal{S}(\bm{\vec{v}})$ added before the linear operation.

$\bm{\tilde{v}}$ is computed following panel (c) of Figure \ref{fig:architecture}, and encodes pose-invariant information from $\vec{v}$, including the vector norms and relative internal angles.
Following again the insight of \cite{Deng2022}, we learn a set of internal axes (in practice, three) within the normalized vector directions via $\bm{W}\bm{\bar{\vec{v}}}$, with $\bm{\bar{\vec{v}}}=\frac{\bm{\vec{v}}}{\lVert\bm{\vec{v}}\rVert}$ and compute their overlaps to $\bar{\vec{v}}$. 
This comprises a transform from the original Cartesian basis to a learned internal basis which rotates equivariantly with the molecule. 
We then concatenate this to the vector norms and pass them through a linear layer, treating them as regular invariant scalars.
The obvious alternative to this `scalarization' scheme is to simply take the euclidean norm of $\bm{\vec{v}}$ and use it as the `scalar representation' of $\bm{\vec{v}}$.
$\bm{\bar{v}}$ lacks invariant information about the relative directions of vectors on the table; $\bm{\tilde{v}}$ should provide a richer picture.

The vector side of the forward pass is analogous to the scalar, but with adjustments to retain equivariance,
\begin{align}\label{eq:vector_forward}
\begin{split}
    \bm{s_g}&=\sigma(\mathrm{N}(\bm{W}^{t}\bm{s}^{t+1} + \bm{b}_s^{t})),\\
    \bm{\vec{v}}_g&=\vec{\sigma}(\vec{N}(\bm{W}^{t}\vec{v}^t))),\\
    \bm{\vec{v}}^{t+1}&=\vec{\bm{v}}^{t} + \bm{s_g}\odot\bm{\vec{v}}_g.
\end{split}
\end{align}
First, we pass information from the scalar track to act as a rescaling factor for vector norms.
Then, we transform the vector features using modified vector normalization, $\vec{N}$, operating only on vector norms, and vector activation $\vec{\sigma}$ (see Fig. \ref{fig:architecture} panel (d)), following the scheme in \cite{Deng2022}.
Finally, the residual is re-added and if required the dimension is adjusted for the next layer.

The graph  model in the encoder follows alternating EMLP and graph convolution layers.
Before node embedding, point positions are first normalized by a radial factor equal to the largest molecular radius in the full dataset.
The initial node embedding sets scalar and vector features for each point in the input cloud,
\begin{align}\label{eq:node_embedding}
\begin{split}
    \bm{x}^0_i &= \bm{W}\left(\bm{E}_{z_i}|\bar{r}_i\right) + \bm{b} \\
    \vec{v}^0_i &= \bm{W}\bar{\vec{r}}_i
\end{split}
\end{align}
with $\vec{r}_i$ the vector from the molecule centroid to atom $i$, $\bar{r}_i$ the norm of $\vec{r}_i$, and $E_{z_i}$ a learned embedding of the atom type $z_i$.

Aggregations during graph convolution and global aggregation are both handled by an augmented softmax operation.
We add to the work of~\cite{li2020deepergcn}, wherein softmax aggregation was shown to smoothly interpolate between the standard `mean' and `max' aggregation operations via a channel-dependent learned temperature, $T_k=1/\beta_k$.
We boost the flexibility of this approach with the addition of a second learned parameter, $b_k$, which adds `sum' to the set of aggregation operations smoothly interpolated by the aggregator, which we call `augmented softmax',
\begin{equation}\label{eq:softmax_aggregation}
    \bm{g}_k = \sum_i^{n_i} \left(\frac{e^{\beta_kx_{ik}}}{\sum_j^{n_i}e^{\beta_kx_{jk}}}+b_k\right) x_{ik},
\end{equation}
where sum is taken over the number of nodes to be aggregated, $n_i$, $x_{ik}$ is the k-th channel feature for node $i$, and $\bm{g}_k$ the aggregated value for channel $k$.

We use this function for the aggregation of both scalars and vectors, with the modification that when aggregating vectors, the norm is substituted for the scalar in the softmax operation
\begin{equation}\label{eq:v_softmax_aggregation}
    \vec{\bm{g}}_k = \sum_i^{n_i} \left(\frac{e^{\beta_k\bar{\bm{v}}_{ik}}}{\sum_j^{n_i}e^{\beta_k\bar{\bm{v}}_{jk}}}+b_k\right) \vec{\bm{v}}_{ik},
\end{equation}
now with $\vec{\bm{v}}_{ik}$ the vector-valued k-th channel feature for node $i$ and $\bar{\bm{v}}_ik$ its Euclidean norm. 

Graph convolution again comprises a scalar and vector track, with scalar messages passed according to
\begin{equation}\label{eq:scalar_conv}
   \bm{s}_i^{t+1} = \bm{s}_i^t+\bm{W}_1Agg_j(\sigma(N(\bm{W}_2\bm{s}_i^t|\bm{W}_3\bm{s}_j^t|\bm{W}_4\bm{e}_{ij}))),
\end{equation}
with $\bm{W}_1$-$\bm{W}_4$ learnable weight matrices, $\bm{e}_{ij}$ the feature vector for the $j\to i$ edge, $j\in r_c$ the nodes $j$ within the cutoff range $r_c$ of node $i$, and $Agg_j$ the augmented softmax aggregation over such nodes.
Vector messages are passed analogously, except we gate against the edge contributions rather than adding to maintain equivariance,
\begin{equation}\label{eq:vector_conv}
   \vec{\bm{v}}_i^{t+1} = \vec{\bm{v}}_i^t+\bm{W}_1Agg^v_j(N(\bm{W}_2\vec{\bm{v}}_i^t + \bm{W}_3\vec{\bm{v}}_j^t) \cdot \bm{W}_4\bm{e}_{ij}),
\end{equation}
with $Agg_j^v$ the vector augmented softmax aggregation.

All codes and training scripts are available at our archived GitHub repository \href{https://github.com/InfluenceFunctional/MXtalTools}{here}.

\section{Mean Deviation Calculation}

For molecule reconstruction, the mean deviation between input and outputs is calculated via a scaffolding and averaging procedure.
For each atom in the input point cloud (molecule), GM output components within 0.5 \r{A} are collected, and a synthetic atom is generated by the weighted sum of their positions.
If the total probability mass of the synthetic atom is above a certain cutoff, in this study, 0.25, we count the input atom as `matched', and compute the distance between the two.

The average over all atoms in a dataset is done simply as $\Delta_{atomwise}=\frac{1}{n}\sum_i^{n_i}\delta_i$, with $\delta_i$ the computed deviation for atom $i$ and $n$ as the number of atoms in the dataset.
The molecule-wise average is done instead by first averaging over the number of atoms in each molecule, and then the number of molecules in the dataset, as $\Delta_{molwise}=\frac{1}{n_m}\sum_m^{n_m}\frac{1}{n_i}\sum_i^{n_i}\delta_i$, for $n_m$ the number of molecules in the dataset, and $n_i$ the number of atoms in molecule $m$.

\section{Regression heads}

Scalar regression and proxy discriminator regression heads are straightforward EMLP models with scalar outputs.
The crystal embedding model is a simple scalar MLP.

For the regression of the QM9s vector properties, we use the vector output of the EMLP model.

For regression of rank 2 tensors (polarizability and quadrupole moment), we adopt the following ansatz: first split the scalar and vector output channels into two equally sized groups, $\bm{s}=\bm{s}_1\oplus\bm{s}_2$, $\vec{\bm{v}}=\vec{\bm{v}}_1\oplus\vec{\bm{v}}_2$, then construct a symmetric tensor via isotropic and anisotropic parts,
\begin{equation}\label{eq:sym_tensor}
    \bm{T}^{(2)}=\sum_i^{n_{out}}\bm{s}_{1,i}\cdot\bm{I}^3 + w_i\frac{1}{2}(\vec{\bm{v}}_{1,i}\otimes\vec{\bm{v}}_{2,i} + \vec{\bm{v}}_{2,i}\otimes\vec{\bm{v}}_{1,i}),
\end{equation}
where $w_i$ is a scalar weight given by a softmax on $s_2$, $\bm{I}^3$ is the identity matrix, and $n_{out}$ is the number of components in the output.

Rank 3 tensors are constructed analogously, but with the ranks increased by one, and the the output channels split into four batches,
\begin{equation}\label{eq:3_tensor}
    \bm{T}^{(3)}=\sum_i^{n_{out}}
    \frac{1}{3}(\bm{I}^3\bm{v}_{4,i} + (\bm{I}^3\bm{v}_{4,i})^{(13)} + (\bm{I}^3\bm{v}_{4,i})^{(12)})
    +(\vec{\bm{v}}_{1,i}\otimes\vec{\bm{v}}_{2,i}\otimes\vec{\bm{v}}_{3,i} + ... + \vec{\bm{v}}_{3,i}\otimes\vec{\bm{v}}_{2,i}\otimes\vec{\bm{v}}_{1,i}),
\end{equation}
with the first sum over the outer products of identity matrices and input vectors with permuted indices, and the sum in the second brackets taken over the six permutations of the indices.
The first term embeds vectors as 3-tensors, replacing the scaled identity matrix, and the second term constructs the full symmetric tensor. 

\section{Model hyperparamters and optimization}

A single training run of one of our autoencoder models on QM9 took 2 days on an NVIDIA RTX8000 GPU. 
A QM9 property regression run took 1-3 hours on the same hardware, and a crystal energy regression takes 4-8 hours.
Over the course of the project, several hundred partial and complete runs were undertaken for the development of the new architecture and loss model, and to identify hyperparameter regimes with fast convergence go high quality loss minima. 
We found by experimentation that different hyperparameters significantly impacted the generalization and training speed of the autoencoder model. 
Here, we give the best parameters found to-date, which were used for the results shown in the main text.

\textbf{Encoder}: GeLU activation, atom type embedding dimension of 32, graph convolution cutoff of 3\r{A}, no dropout, node embedding dimension of 512, bottleneck dimension of 64, 2 fully-connected layers per EMLP block, message dimension of 64, graph-wise layer normalization on scalars but not vectors, 2 convolutional layers, and 50 Gaussian radial basis functions.

\textbf{Decoder}: GeLU activation, no dropout, 64 nodes per graph, node dimension of 128, 4 convolutional layers, and graph-wise layer normalization on scalars but not vectors.

\textbf{Autoencoder Optimization}: AdamW~\cite{loshchilov2017decoupled} optimizer with weight decay of 0.005, initial learning rate of 5.0E-5, ramping to 1.0E-3, then down to 1.0E-6. Batch size starting at 10 and ramping until GPU memory is maxed out at 167. The Gaussian width is initialized at 2, and reduced by 1\% when the reconstruction test loss reaches 0.15, until the mean sample self-overlap reaches 1.0E-3.

\textbf{Embedding Regressor}: GeLU activation, no dropout or normalization, hidden dimension of 256, and 4 fully-connected layers. 

\textbf{Regressor Optimization}: AdamW optimizer with weight decay of 0.1, initial learning rate of 1.0E-4, ramping to 5.0E-4, then down to 5.0E-6. Batch size starting at 50 and ramping to 2000. 

\textbf{Crystal Energy Model}: GeLU activation, hidden dimension of 512, and 32 fully-connected layers. Layer normalization and 50\% dropout probability. 

\textbf{Energy Model Optimization}: AdamW optimizer with weight decay of 0.1, initial learning rate of 1.0E-4, ramping to 5.0E-4, then down to 5.0E-6. Batch size starting at 50 and ramping to 10000. 

\section{Molecule property prediction}

\begin{figure*}
\centering
    \includegraphics[trim=0 0 0 40, clip, width=\textwidth]{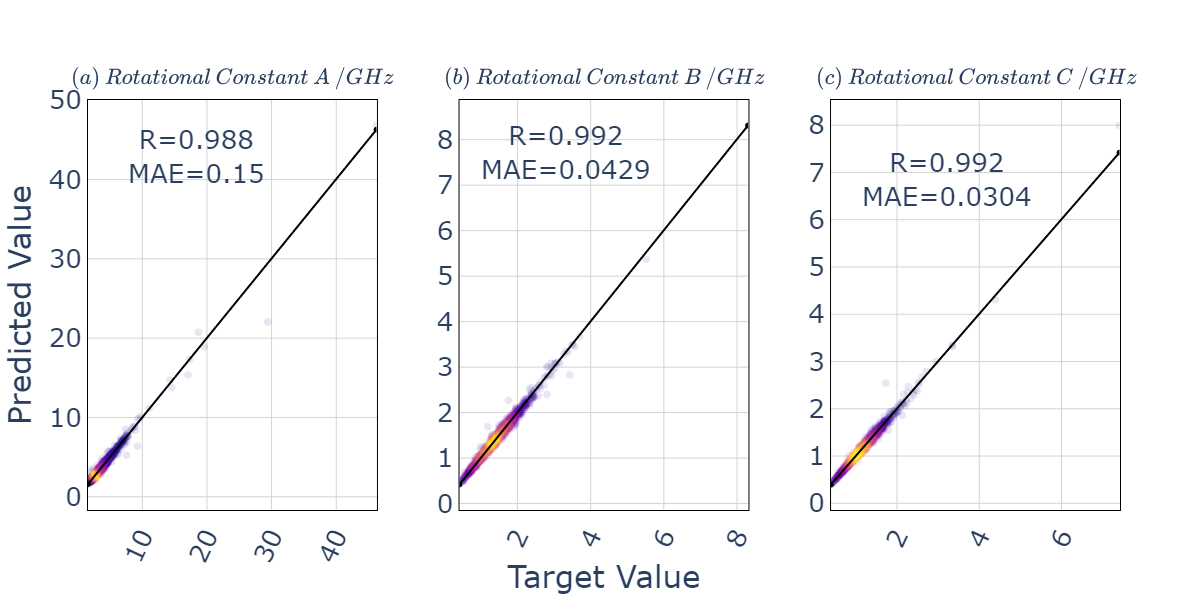}
    \caption{\label{fig:er_0} Regression results for rotational spectroscopic constants.
    }
\end{figure*}

\begin{figure*}
\centering
    \includegraphics[trim=0 0 0 40, clip, width=\textwidth]{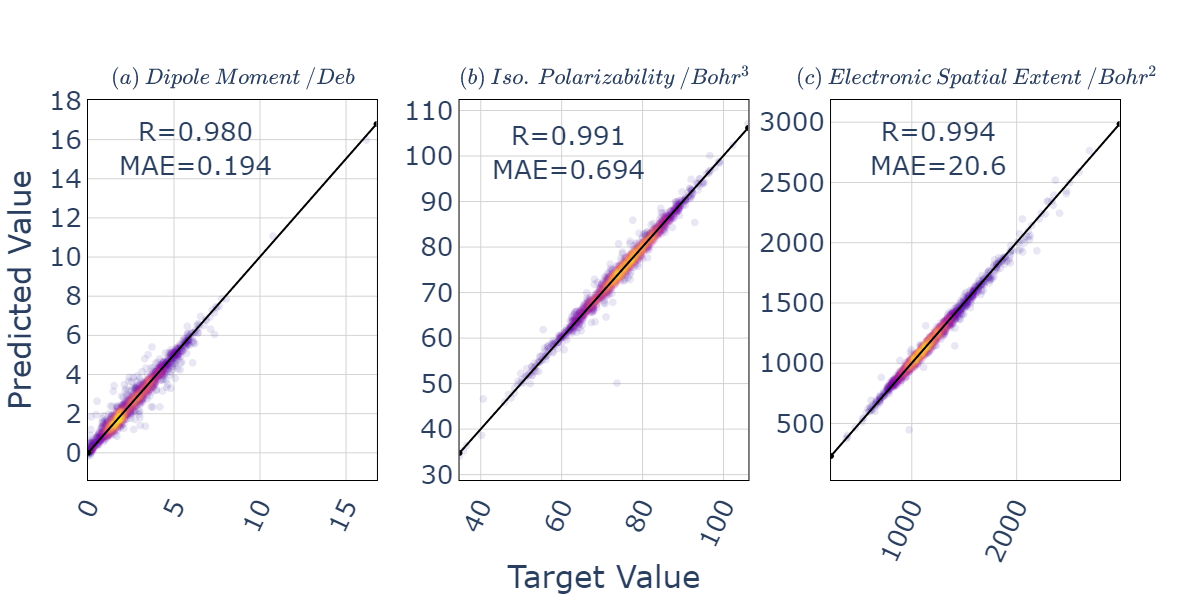}
    \caption{\label{fig:er_1} Regression results for electronic properties.
    }
\end{figure*}

\begin{figure*}
\centering
    \includegraphics[trim=0 0 0 40, clip, width=\textwidth]{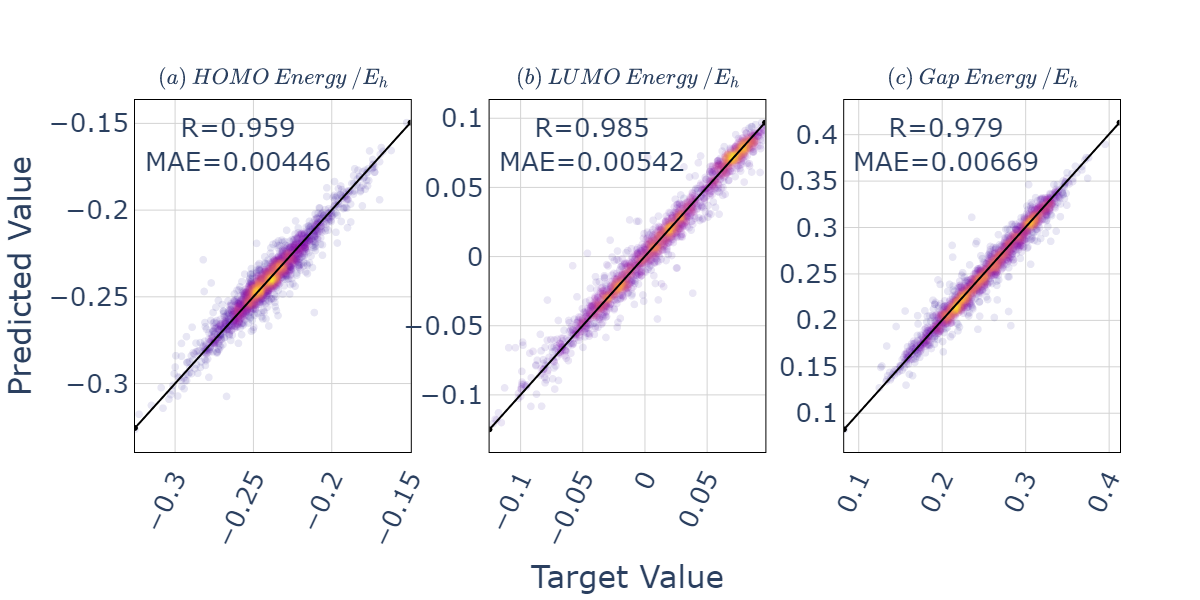}
    \caption{\label{fig:er_2} Regression results for molecular orbital energies.
    }
\end{figure*}

\begin{figure*}
\centering
    \includegraphics[trim=0 0 0 40, clip, width=\textwidth]{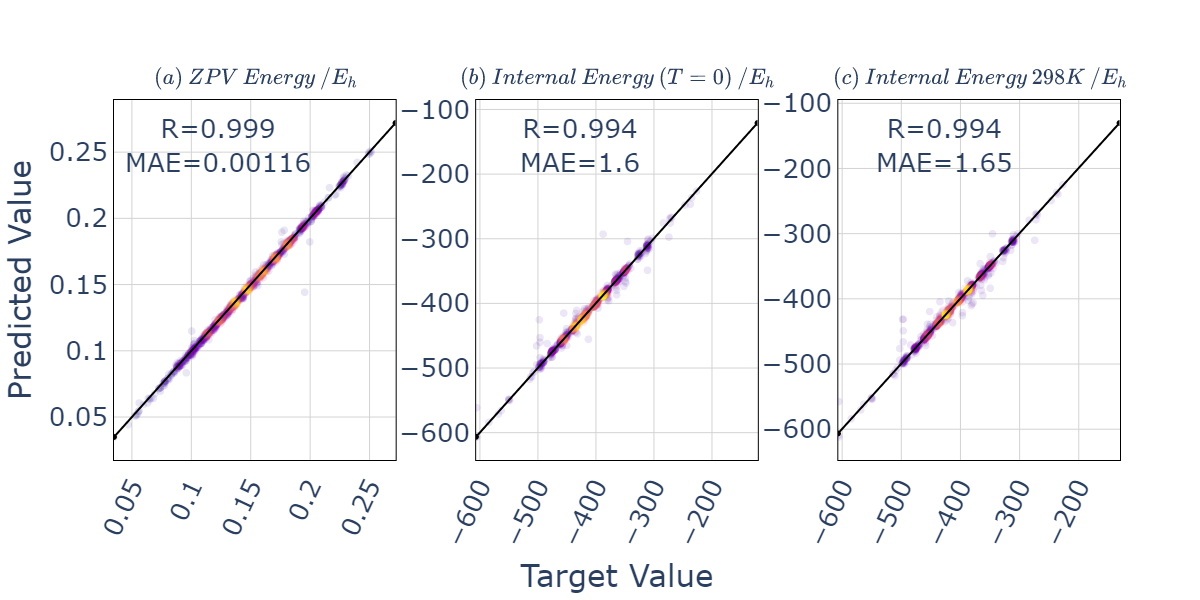}
    \caption{\label{fig:er_3} Regression results for zero point vibrational and internal energies.
    }
\end{figure*}

\begin{figure*}
\centering
    \includegraphics[trim=0 0 0 40, clip, width=\textwidth]{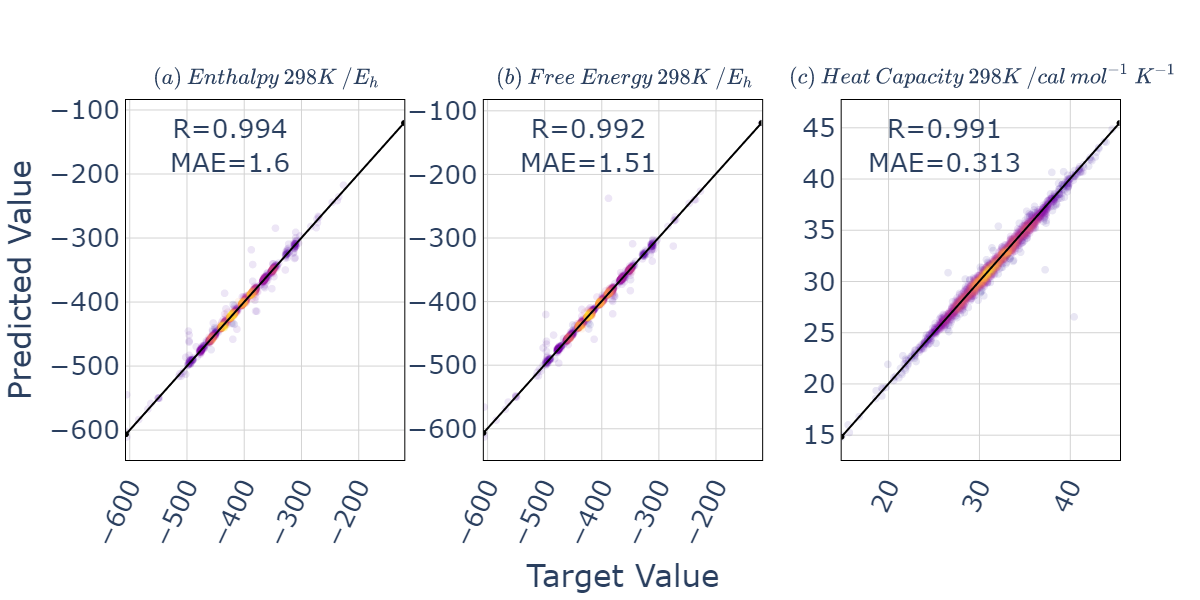}
    \caption{\label{fig:er_4} Regression results for enthalpy, free energy, and heat capacity.
    }
\end{figure*}

\begin{figure*}
\centering
    \includegraphics[trim=0 0 0 40, clip, width=\textwidth]{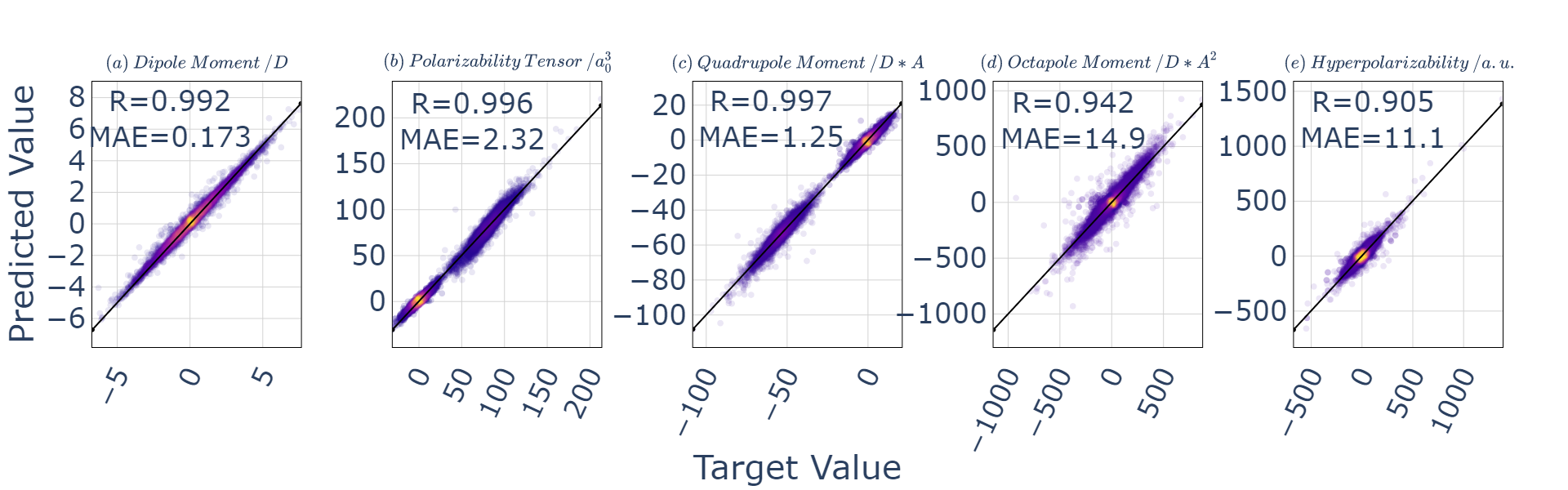}
    \caption{\label{fig:er_5} Regression results for QM9s vector and tensor properties.
    }
\end{figure*}

For reference, we provide in Figures~\ref{fig:er_0}-~\ref{fig:er_5} a more detailed view of the regression results for the QM9/QM9s molecular properties.

\section{Crystal Parameterization}

\begin{figure*}
\centering
    \includegraphics[trim=0 120 0 80, clip, width=\textwidth]{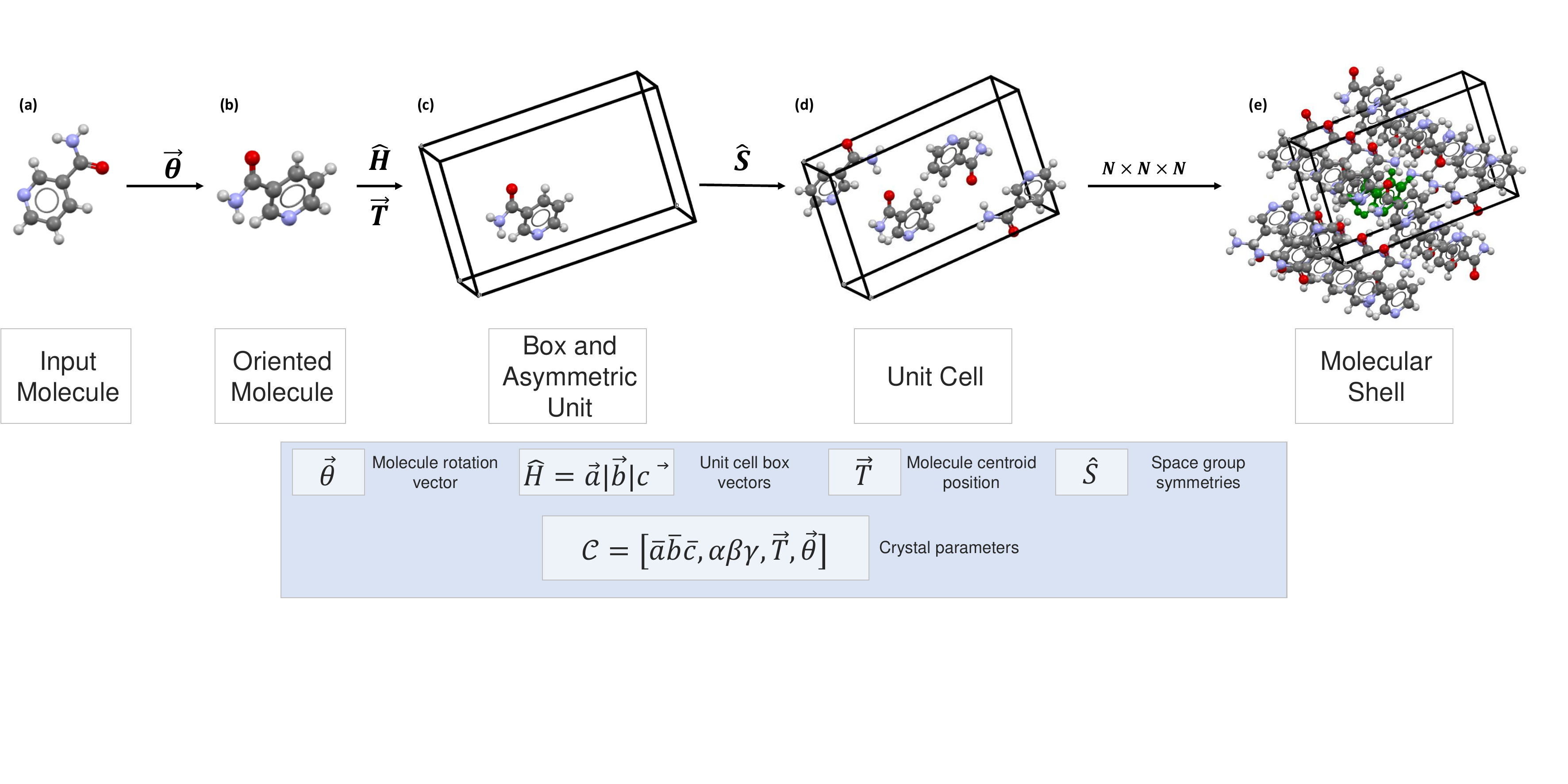}
    \caption{\label{fig:crystal_build} Molecular crystal parameterization, construction, and scoring.
    The molecule comprising the asymmetric unit is taken in a given pose, reoriented by an axis-angle rotation vector $\mathbf{\vec{\theta}}$ and placed at position $\mathbf{\vec{T}}$ in the unit cell defined by $\mathbf{\hat{H}}$.
    The symmetry operations $\mathbf{\hat{S}}$ are applied depending on the relevant space group, and a cluster of molecules within the intermolecular interaction radius of the asymmetric unit is added.
    The intermolecular energy is computed on edges between the asymmetric unit, highlighted in green in panel (e), within a range of 6 \r{A}.
    We show here a molecule in space group 14 ($P2_1/c$), for illustration. 
    Actual training runs were all done in space group 1 ($P$1), where the asymmetric unit is identical to the unit cell.
    }
\end{figure*}

Molecular crystals with $Z'=1$ (one molecule per asymmetric unit) can be completely characterized by the molecule point cloud, the space group, and the 12 crystal dimensions: cell vector lengths ($\bar{a}$,$\bar{b}$,$\bar{c}$), internal angles ($\alpha$, $\beta$, $\gamma$), the position of the molecule centroid in fractional coordinates $\vec{T}=(u, v, w)$, and the molecule orientation defined against a standardized orientation, defined by a rotation vector $\vec(\theta)=(x, y, z)$.
The chosen standard is to align the molecule principal eigenvectors to the cartesian axes.

Lennard-Jones potentials for each molecular crystal are computed over intermolecular edges as
\begin{equation}
    E_{LJ} = 4\sum_{ij} \left(\frac{\sigma_{ij}}{r_{ij}^{12}}-\frac{\sigma_{ij}}{r_{ij}^{6}}\right),
\end{equation}
for $\sigma_{ij}$ the sum of atoms $i$ and $j$'s van der Waals radii.

\end{document}